\documentclass{article} %
\usepackage{graphicx}
\graphicspath{{figures/}}
\usepackage{authblk}
\DeclareUnicodeCharacter{202F}{ }
\graphicspath{ {./figures/} }
\usepackage[colorlinks=false]{hyperref}
\usepackage{float}
\usepackage{mathtools}
\usepackage{amsmath}
\usepackage{amssymb}
\usepackage{listings}
\usepackage[table]{xcolor}
\usepackage{colortbl}
\definecolor{softgreen}{RGB}{0, 153, 0} 

\usepackage{xcolor}
\usepackage{booktabs}
\usepackage{array}
\usepackage{tikz} 
\usepackage{tabularx}
\usetikzlibrary{positioning, shapes.multipart}
\usepackage{multirow}
\usepackage{makecell}
\usepackage{tcolorbox}
\usepackage{verbatim}
\usepackage{geometry}
\usepackage{tablefootnote} %
\usepackage{arydshln}
\definecolor{darkbluegrey}{rgb}{0.2, 0.3, 0.5}
\usepackage{siunitx}
\usepackage[
  backend=biber,
  backref=true
  style=numeric,
  sorting=none,
  maxbibnames=6,  %
  minbibnames=3
]{biblatex}
\usepackage[colorlinks=false]{hyperref}
\usepackage{cleveref}  
\usetikzlibrary{arrows.meta}
\usetikzlibrary{calc}

\title{An Agentic Approach to Generating XAI-Narratives}

\author[1]{Yifan He}
\author[1]{David Martens}
\affil[1]{University of Antwerp, Belgium}
\date{}

\begin{document}
\maketitle

\begin{abstract}
Explainable AI (XAI) research has experienced substantial growth in recent years. Existing XAI methods, however, have been criticized for being technical and expert-oriented, motivating the development of more interpretable and accessible explanations. In response, large language model (LLM)-generated XAI narratives have been proposed as a promising approach for translating post-hoc explanations into more accessible, natural-language explanations. In this work, we propose a multi-agent framework for XAI narrative generation and refinement. The framework comprises the \textit{Narrator}, which generates and revises narratives based on feedback from multiple \textit{Critic Agents} on faithfulness and coherence metrics, thereby enabling narrative improvement through iteration. We design five agentic systems (\textit{Basic Design, Critic Design, Critic-Rule Design, Coherent Design}, and \textit{Coherent-Rule Design}) and systematically evaluate their effectiveness across five LLMs on five tabular datasets. Results validate that the \textit{Basic Design}, the \textit{Critic Design}, and the \textit{Critic-Rule Design} are effective in improving the faithfulness of narratives across all LLMs. Claude-4.5-Sonnet on \textit{Basic Design} performs best, reducing the number of unfaithful narratives by 90\% after three rounds of iteration. To address recurrent issues, we further introduce an ensemble strategy based on majority voting. This approach consistently enhances performance for four LLMs, except for DeepSeek-V3.2-Exp. These findings highlight the potential of agentic systems to produce faithful and coherent XAI narratives. 

\bigskip

\end{abstract}

\section{Introduction}
With the rapid development of black-box machine learning models, growing concern about transparency and accountability, and increasing regulatory and compliance requirements in high-stakes domains, explainable AI research has seen a substantial rise in recent years~\cite{long2025explainableailatestadvancements,BARREDOARRIETA202082}.  However, as early as 2017, scientists had already argued that predominant explainable AI methods are too technical-driven, which pose limitations on lay-users' understanding~\cite{miller2017explainableaibewareinmates}. SHAP, as one of the most widely used approaches among existing XAI techniques~\cite{molnar2025}, is no exception to this issue~\cite{martens2024tellstorynarrativedrivenxai,Liao_2020}. As a feature-attribution explanation tool, SHAP quantifies feature contribution to a model’s prediction at an abstract numerical level, hindering users’ understanding of model behavior. 

Recently, LLM-generated XAI narratives have been proposed as a promising solution that transforms technical XAI explanations into human-readable narratives~\cite{martens2024tellstorynarrativedrivenxai}. This is based on science communication theory, which attributes narratives as persuasive, accessible, and effective in engaging non-experts~\cite{doi:10.1073/pnas.1320645111}. More importantly, advances in LLMs enable this approach, transforming XAI explanations into narratives in an automatable, scalable way with little to no human input~\cite{Howgoodismystory}. Some studies directly use LLMs to generate post-hoc explanations~\cite{bhattacharjee2024llmguidedcausalexplainabilityblackbox,kroeger2023are}. However, such approaches have been criticized for introducing more uncontrollable faithfulness issues to the prediction model~\cite{zytek2024explingoexplainingaipredictions}.  In this study, we primarily review work that transforms post-hoc XAI outputs into natural-language narratives using LLMs.  

By different data types, prior studies explore narrative generation, including graphs~\cite{cedro2025graphxainnarrativesexplaingraph,giorgi2025naturallanguagecounterfactualexplanations}, images~\cite{martens2024tellstorynarrativedrivenxai,wojciechowski2025faithfulplausiblenaturallanguage}, and tabular datasets~\cite{martens2024tellstorynarrativedrivenxai,deoliveira2023unveilingpotentialcounterfactualsexplanations,zytek2024llmsxaifuturedirections,zytek2024explingoexplainingaipredictions, giorgi2025enhancingxainarrativesmultinarrative}. Specifically, Martens et al.~\cite{martens2024tellstorynarrativedrivenxai} investigated both SHAP-based narratives for tabular data and counterfactual-based narratives for images. Zytek et al.~\cite{zytek2024explingoexplainingaipredictions} concentrated on SHAP narrative generation by introducing \textit{Narrator} and \textit{Grader}, where the former generates SHAP narratives and the latter evaluates them using an LLM-as-a-judge approach. In addition, Giorgi et al.~\cite{giorgi2025enhancingxainarrativesmultinarrative} generate narratives for counterfactual explanations.

In 2024, both industry commentary and academic discussions highlighted Agentic AI as a promising research direction~\cite{Belcic2025-vr,xi2023risepotentiallargelanguage}. This trend introduces an alternative paradigm for XAI narrative research. Recent studies have explored agentic system approaches for XAI, some of which do not rely on existing post-hoc explanation methods~\cite{bandara2025responsibleexplainableaiagents}, while others do, yet remains limited~\cite{10.1007/978-3-032-06078-5_9,yamaguchi2025agenticexplainableartificialintelligence}. Most of them focus on an interactive, natural-language explanations through conversational interface~\cite{slack2023talktomodelexplainingmachinelearning,Nguyen_2023,shen2023convxaideliveringheterogeneousai,10.1007/978-3-030-30391-4_5,10.1007/978-3-032-06078-5_9}. For example, Slack et al.~\cite{slack2023talktomodelexplainingmachinelearning} proposed TalkToModel, an interactive dialogue system that allows users to ask further questions on explanations through conversations. A closely related study is presented by Serafim et al.~\cite{10.1007/978-3-032-06078-5_9}, who proposed MAINLE, a four-agent architecture for generating XAI narratives. This multi-agent architecture automate the full process of narrative generation, yet without an agent designed to identify the possible issues and refine them. 

To summarize, while several studies have successfully generated XAI narratives using LLMs, only a small subset uses an agentic approach. Among these, most focus primarily on user interaction and conversational deployment, and they typically experiment with a single LLM without comparing performance across models. Moreover, in existing studies employing multi-agent architectures, automated and real-time refinement is not incorporated; instead, all assessments occur only after generation, via human feedback.

Another central challenge in this field lies in evaluating natural-language narratives. In the context of the intersection between LLMs and XAI, the survey by Bilal et al.~\cite{bilal2025llmsexplainableaicomprehensive} categorizes the evaluation of LLM-generated explanations into two broad categories: (1) qualitative evaluations, such as comprehensibility and controllability, and (2) quantitative evaluations, such as faithfulness and plausibility. Focusing more specifically on LLM-generated XAI narratives, a systematic review by Silvestri et al.~\cite{LLMXAINarrative} argues that XAI narratives, due to its linguistic nature, are expected to not only accurately convey information from the underlying explainer, but also ensure fluency, grammatical correctness, and other essential linguistic quality characteristics. Accordingly, three evaluation dimensions are identified as mainstream metrics: faithfulness to the explainer, linguistic realization (such as fluency), and human utility (such as cognitive fit). We observe that current studies choose different dimensions depending on their research objectives, often covering more than one. For example, some assess faithfulness~\cite{zytek2024llmsxaifuturedirections, zytek2024explingoexplainingaipredictions, 10.1007/978-3-032-06078-5_9,Howgoodismystory}, linguistic quality~\cite{zytek2024llmsxaifuturedirections,zytek2024explingoexplainingaipredictions,giorgi2025enhancingxainarrativesmultinarrative,10.1007/978-3-032-06078-5_9,Howgoodismystory}, and others more on human utility, such as comprehensibility~\cite{slack2023talktomodelexplainingmachinelearning,martens2024tellstorynarrativedrivenxai,giorgi2025enhancingxainarrativesmultinarrative, 10.1007/978-3-032-06078-5_9}. With respect to evaluation methods, researchers typically adopt either automatic approaches~\cite{Howgoodismystory}, human-centered methods such as user surveys~\cite{zytek2024llmsxaifuturedirections,slack2023talktomodelexplainingmachinelearning,martens2024tellstorynarrativedrivenxai,giorgi2025enhancingxainarrativesmultinarrative,10.1007/978-3-032-06078-5_9}, or LLM-based approaches, especially the emerging paradigm of LLM-as-a-judge~\cite{zytek2024explingoexplainingaipredictions,10.1007/978-3-032-06078-5_9}, in which an LLM grades a narrative according to a predefined rubric. In the next section, we will discuss two evaluation metrics — faithfulness and coherence, the reasons we finally use them in this study, and why we use coherence as an umbrella concept when evaluating linguistic quality. 

Faithfulness is one of the foundational metrics of a high-quality XAI narrative. If a narrative contains information that conflicts with the output of the underlying explainer, it can hardly be considered reliable or useful~\cite{LLMXAINarrative}. Faithfulness measures the degree of alignment between the information conveyed in the narrative and the XAI method's explanation~\cite{LLMXAINarrative}, with the assumption that the explanation itself is fully faithful to the prediction model~\cite{Howgoodismystory}. For example, when the XAI method is SHAP, a narrative that mentions a feature not present in the SHAP table constitutes a faithfulness-related issue. Based on a Likert scale or pre-defined rubric for scoring, some studies leverage either LLM-as-a-judge~\cite{zytek2024explingoexplainingaipredictions, 10.1007/978-3-032-06078-5_9} or a user study~\cite{zytek2024llmsxaifuturedirections,giorgi2025naturallanguagecounterfactualexplanations, 10.1007/978-3-032-06078-5_9} to grade the faithfulness of a narrative. Although few, other studies implement an automatic assessment of faithfulness. For instance, Ichmoukhamedov et al.~\cite{Howgoodismystory} first extract faithfulness-related information from the narrative through LLMs. They use three downstream metrics, namely \textit{Rank Agreement}, \textit{Sign Agreement}, and \textit{Value Agreement}, to compare against the ground-truth information in the SHAP input. This type of quantitative, automated evaluation of faithfulness is particularly useful and necessary for studies that require an uninterrupted process, such as an agentic approach. 

On the linguistic quality side, fluency is an important component. It measures the grammatical correctness, the natural flow, and sometimes even the readability level~\cite{LLMXAINarrative, shen2023largelanguagemodelshumanlevel}. Some paper specially distinguish fluency from coherence, by arguing that fluency is at the sentence level while coherence is the collective quality of all
sentences~\cite{shen2023largelanguagemodelshumanlevel}. In contrast, other works use several terms, such as fluency, conciseness, coherence, among others, in a collective or overlapping manner, to assess the overall linguistic quality of a narrative \cite{giorgi2025enhancingxainarrativesmultinarrative, yamaguchi2025agenticexplainableartificialintelligence, zytek2024explingoexplainingaipredictions,zytek2024llmsxaifuturedirections}. Beyond surface-level linguistic quality, plausibility is regarded as an important evaluation dimension in the XAI explanation field. Given that LLMs can introduce assumptions regarding feature influence on prediction outcomes, plausibility evaluates whether such explanations are logically consistent and aligned with domain knowledge. Ichmoukhamedov et
al. \cite{Howgoodismystory} evaluate plausibility using averaged perplexity scores computed over injected feature assumptions, based on the assumption that lower perplexity corresponds to smoother and more plausible explanations. However, their results indicate that perplexity is an ineffective plausibility proxy when assumptions are embedded within full narrative paragraphs instead of in isolation. Another study uses perplexity to evaluate fluency and coherence \cite{shirvanimahdavi2025rule2textframeworkgeneratingevaluating}, being a measure that complements correctness (faithfulness). All in all, given the complex overlap among several concepts and the limitations of perplexity as a plausibility proxy, we adopt \textit{coherence} as an umbrella concept in this study. This concept encompasses fluency, grammatical correctness, and logical flow to represent the overall linguistic quality of XAI narratives. 

\textbf{Present Work } 
As discussed earlier, although LLM-generated XAI narratives have been explored in prior work, the use of agentic approaches in this context remains underexplored. Existing agentic systems are typically tested on a single LLM, without further comparisons across different models' performance, and tend to emphasize user conversational interfaces as their primary objectives. At the same time, the evaluation of XAI narratives is largely post hoc, with no integration of real-time feedback into the generation process. Researchers also rely mainly on user surveys or LLM-based graders, with automated methods insufficiently explored. 

This work proposes a multi-agent architecture for generating and improving SHAP-based narratives\footnote{All the code, experiments, and the results can be found in this GitHub link: https://github.com/ADMAntwerp/SHAPnarrative-agents}. Specifically, this study:

\begin{enumerate}
    \item Proposes a fully-automated multi-agent framework capable of both generating and refining SHAP-based narratives.
    \item Integrates quantitative and qualitative evaluation processes into the iterative loop, enabling real-time refinement on two key metrics: the faithfulness and coherence of XAI narratives.
    \item Validates the effectiveness of the agentic approach across five LLMs (GPT-5, Claude Sonnet 4.5, DeepSeek-V3.2-Exp, Mistral Medium 3.1, and Llama 3.3 70B Instruct), five agentic system designs (Basic Design, Critic Design, Critic-Rule Design, Coherent Design, and Coherent-Rule Design), and two architectures (original and ensemble).
    \item Examine the effectiveness conditions of the agentic approach across several key concepts, including the maximum number of refinement rounds, the number of features in the narrative, the quality of baseline narratives, and the no-baseline setting.
\end{enumerate}

\begin{figure}[hbt!]
    \centering
    \includegraphics[width=0.8\textwidth]{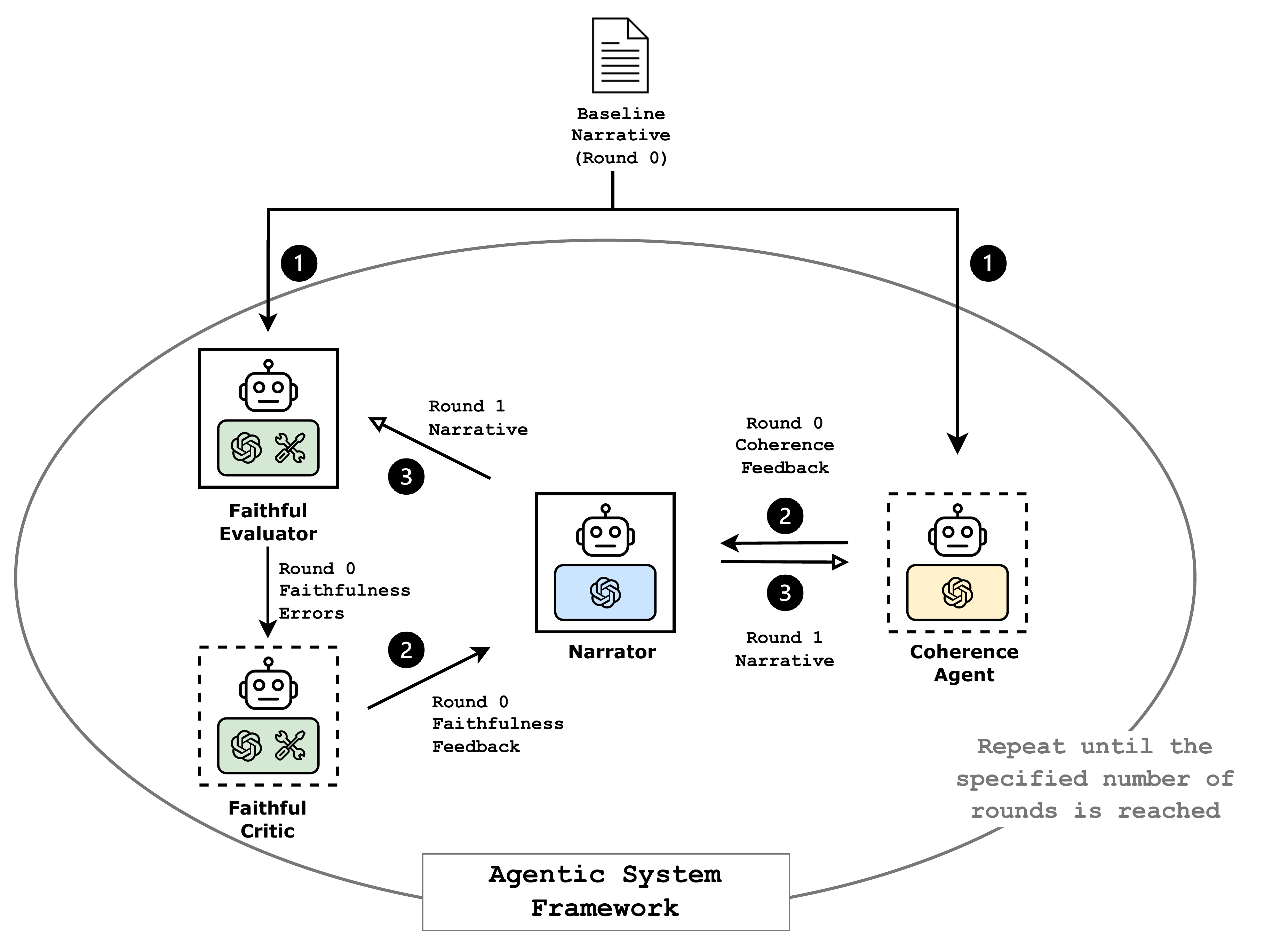}
    \caption{The baseline narratives are fed into the \textit{Faithful Evaluator} and the \textit{Coherence Agent} in the beginning. Based on the \textit{Faithful Evaluator}'s output, the \textit{Faithful Critic} offers revision suggestions to the \textit{Narrator}. On the other side, the \textit{Coherence Agent} identifies coherence-related issues and provides revision suggestions to the \textit{Narrator}. The \textit{Narrator} then generates an updated version of the narrative and continues the iteration until the stopping criterion is met. }
    \label{schematic}
\end{figure}

The agentic system schematic is shown in Figure~\ref{schematic}. First, the baseline narratives (Round 0\footnote{We start the round from ``0" to better signal the generated narratives before any feedback or refinement. It is not part of the iterative improvement loop yet. The first set of narratives generated by the \textit{Narrator} is called ``round-1 narratives". }) are fed into the \textit{Faithful Evaluator} and the \textit{Coherence Agent}. The \textit{Faithful Evaluator} extracts feature information from the narrative, compares it with the original SHAP input table, and outputs the existing errors, including ``rank", ``sign", and ``value" errors. Based on the \textit{Faithful Evaluator}'s work, the \textit{Faithful Critic} offers directional and actionable feedback about how to fix those errors. On the other side, the \textit{Coherence Agent} identifies coherence-related issues, such as awkward transitions and logical inconsistencies, and outputs modification suggestions. Subsequently, the \textit{Narrator} receives both feedback and generates an updated version of the narrative (Round 1). This narrative is then fed back into the \textit{Faithful Evaluator} and the \textit{Coherence Agent} again, as the start of round 1. The iterative process continues until a stopping criterion is met, either when the maximum number of rounds is reached or when the narrative is judged 100\% faithful. Depending on the design of the agentic system, not all agents are included in every version (agents shown in dashed boxes are optional); however, the \textit{Narrator} and the \textit{Faithful Evaluator} are always present as fixed components.

\section{Methodology}

\subsection{Agent Players and Roles}
The purpose of this task is to autonomously generate and refine a narrative, and output an upgraded version for a given instance. In this proposed multi-agent system, different agents interact with each other to complete this task, with each agent focusing on a specific one, either generating a narrative or giving feedback based on their evaluation goal. Four agents are present in this system: \textit{Narrator}, \textit{Faithful Evaluator}, \textit{Faithful Critic}, and \textit{Coherence Agent}. We also design a non-LLM variant for the \textit{Faithful Critic}, denoted as the \textit{Faithful Critic (Rule)}. Different sets of agents are employed in five agentic system designs for further comparison (Section~\ref{architecture}).  In this section, we describe each agent's responsibilities, inputs and outputs, and preparatory steps. In the following section, we present a schematic illustrating how these agents interact. 

\textbf{Prompt }The main content of the \textit{Narrator} and the \textit{Faithful Evaluator}'s prompts are largely based on \cite{Howgoodismystory}\footnote{The prompt content can be found in each agent’s script:\url{https://github.com/ADMAntwerp/SHAPnarrative-agents/tree/clean/shapnarrative_agents/agents}}. We refer to~\cite{zytek2024explingoexplainingaipredictions} for the overall prompt structure, and also follow the prompt design principles outlined in~\cite{bsharat2024principledinstructionsneedquestioning} to refine the wording. The prompt structure for all agents is consistent: it starts with the \textit{context}, which includes background information and their own roles and tasks, then the \textit{input} text (e.g., the narrative and feedback from that round), and ends with the output requirements and additional guidelines. We will further discuss some details in the following for each agent, while the full content of the prompts is provided in Appendix~\ref{baseprompt}.

\textbf{Preparation} Some necessary information should be prepared before the introduction of the agentic system. We implement a Class named ``prompt" that is equipped with several built-in methods to prepare them all. Given the dataset information and a single instance, it predicts the target class, probability scores, and the SHAP input table. The SHAP table is derived from the SHAP tree explainer\footnote{https://github.com/shap/shap}, with every row containing a feature name, a SHAP value, a feature value, the averaged feature value of all instances, and a feature description. The feature ranking in the SHAP table follows the absolute SHAP values, with the most important feature (either positive or negative influence) listed in the first row. All of the above information, together with explicit generation guidelines (e.g., format and content constraints), is formulated into a \textit{base prompt}, which is subsequently used by the \textit{Narrator}. Following the \textit{base prompt}, the generated narrative should start with a clear prediction of results, with the explanation of the four most important features and their influence, ending with a summary. 

The formulation of the \textit{base prompt} content is adapted from the method proposed in \cite{Howgoodismystory} (long version) \footnote{ Their work evaluates two prompt variants (a long version and a short version) where the long version includes more detailed explanations of SHAP, task descriptions, and related context. The results demonstrate that the long prompt consistently outperforms the short one, leading us to adopt the long version in our study.}.They allow control over the maximum narrative length and the number of features included in the explanation. They ultimately limit narratives to at most 10 sentences and include only the four most important features (n=4) to ensure brevity. We thus directly follow the same settings in our main experiments.  

\begin{figure}[htb]
\centering
\scalebox{0.55}{
\input{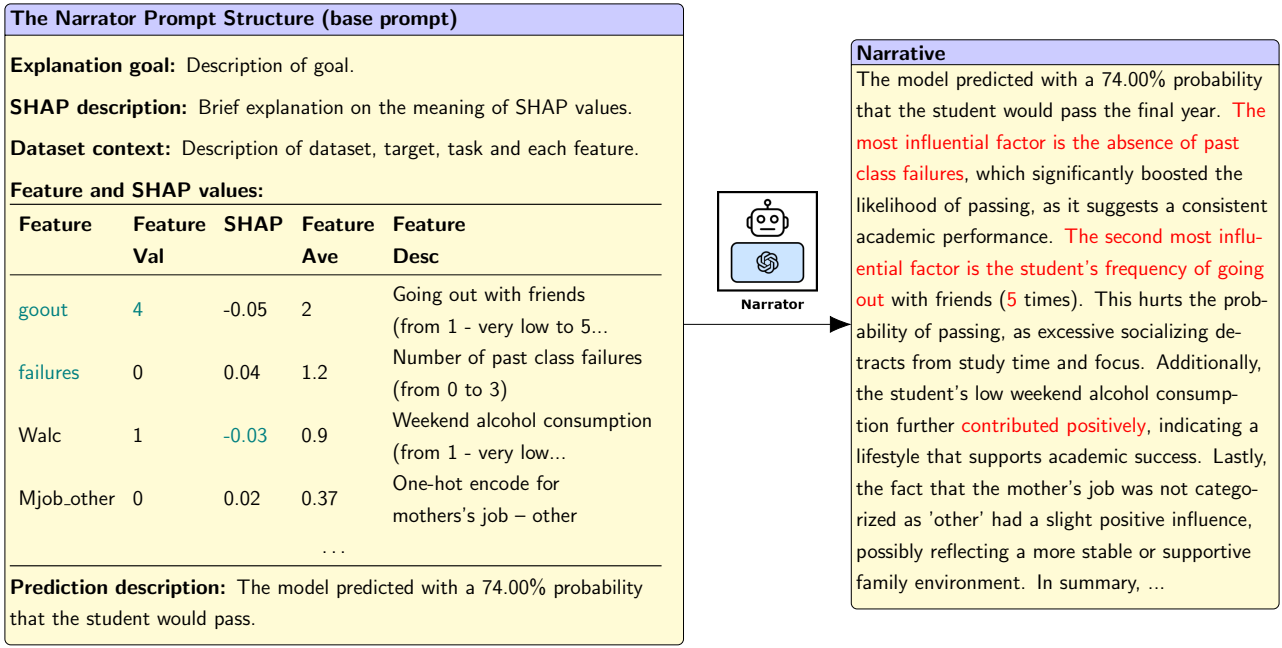}}
\caption{\textbf{Left:} This is an example of the base prompt for the \textit{Narrator}. The instance is from the \textit{Student} dataset. \textbf{Right:} The \textit{Narrator} generates a narrative based on the given SHAP input. However, in this example, it injects multiple faithfulness-related issues, including the wrong feature value on ``goout'' (being 5 times instead of 4 times in the given SHAP input), the wrong feature sign on ``Walc'' (being ``positive" instead of ``negative"), and the interchanged feature importance ranking between ``goout'' and ``failures''. These issues are marked in red.}
\label{NA}
\end{figure}

\textbf{The Narrator} The \textit{Narrator} is designed to generate and revise narratives. In round 0, the baseline narratives are directly fed into the system, as shown in Figure~\ref{schematic}. Under the no-baseline setting, the \textit{Narrator} responds to the \textit{base prompt} and generates the round-0 narrative. In subsequent rounds, the \textit{Narrator} receives a new prompt that combines the \textit{base prompt}, the last round narrative, and feedback provided by the other agents, such as the \textit{Faithful Evaluator}, and the \textit{Coherence Agent} (if present). In response, it revises the narrative and generates an updated version. An example of the Narrator's response to the prompt is shown in Figure~\ref{NA}.

\begin{figure}[hb]
\centering
\scalebox{0.55}{
\input{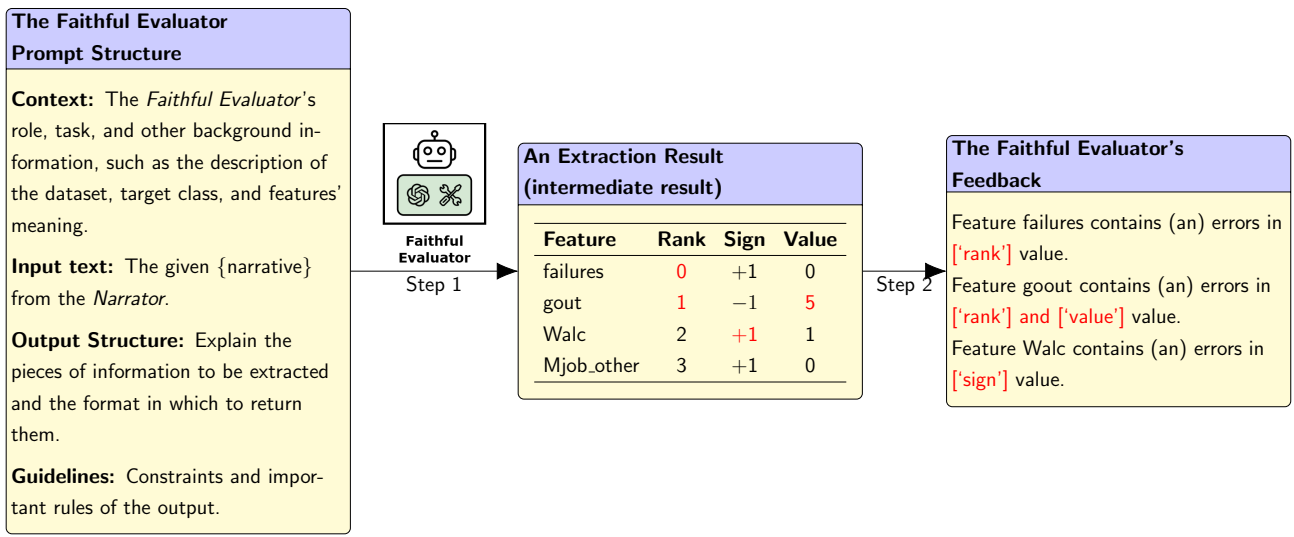}}
\caption{\textbf{Left:} The \textit{Faithful Evaluator}'s prompt that we use to extract information from a narrative. The narrative from Figure~\ref{NA} is fed into the prompt. \textbf{Middle:} The \textit{Faithful Evaluator} extracts the information into a dictionary, including the rank, sign, and value information for each feature. This is an intermediate output, which can be reused by the \textit{Faithful Critic}. \textbf{Right:} The output from the \textit{Faithful Evaluator}. The errors contained in the narrative are captured and reported by the \textit{Faithful Evaluator}, shown in red. }
\label{FE}
\end{figure}

\textbf{The Faithful Evaluator} The \textit{Faithful Evaluator} is designed to output faithfulness-related errors that exist. In the first step, we use a prompt that allows an LLM to extract the related information from a narrative into a dictionary. The prompt contains the \textit{Faithful Evaluator}'s role and task, the narrative to be revised, the necessary background information (e.g., description of the dataset and features), and explicit output structure and guidelines. The expected output is a dictionary in which feature names serve as keys, each with three inner keys: ``rank", ``sign", and ``value", and their associated values.  Given all input features $\mathcal{F}$, for each feature $f_{j} \in \mathcal{F} : j \in \lbrace{0,...,n-1 \rbrace}$ (Figure~\ref{FE}, step 1):

\begin{itemize}
    \item \textbf{Rank}: natural numbers $ 0 \leq r_j \leq n-1$ represents the importance rank of feature $f_j$ as implied in the narrative. $r=0$ represents the most important feature, which has the largest absolute SHAP values, regardless of positive or negative influence. As stated, we adopt the setting from~\cite{Howgoodismystory} and include only the four most important features from the SHAP, therefore n=4 in the main experiments of this study. Typically, the extracted $r_j$ follows the appearance order of a feature in a narrative, hence $r_j=j$. In rare cases, they are different due to inappropriate description terms, which are discussed in Section \ref{categorization}.
    \item \textbf{Sign:} $s_j \in \lbrace{ -1,1\rbrace}$. The feature $f_j$ represents either a negative or positive influence on the prediction, respectively.
    \item \textbf{Value:} $v_j \in \mathbb{R} \cup \lbrace \phi \rbrace$. The value field represents the feature’s actual value as described in the narrative. For a discrete feature, the feature value can either be 0 or 1 due to the dummy encoding. For a continuous feature, the feature value can be any real number. The extraction of a feature value is also flexible: if a value is not explicitly stated in the narrative, the LLM is allowed to not extract any feature value, thus returning a null value $\phi$. 
\end{itemize}

Afterwards, the \textit{Faithful Evaluator} compares the extracted values under each key against the original SHAP input and reports any mismatches as faithfulness errors (Figure~\ref{FE}, step 2). Without using an LLM at this stage, a simple rule-based method compares the differences and outputs a Boolean value indicating whether a mismatch is detected. The final feedback provided by the \textit{Faithful Evaluator} is in a fixed format: ``Feature \{feature\_name\} contains (an) errors in {[`rank']/[`sign']/[`value']} value.'', depending on the type of error. When there is no mistake in the narrative, the output is:  ``After checking, the narrative is 100\% faithful to the SHAP table.''

\begin{figure}[htb]
\centering
\scalebox{0.55}{
\input{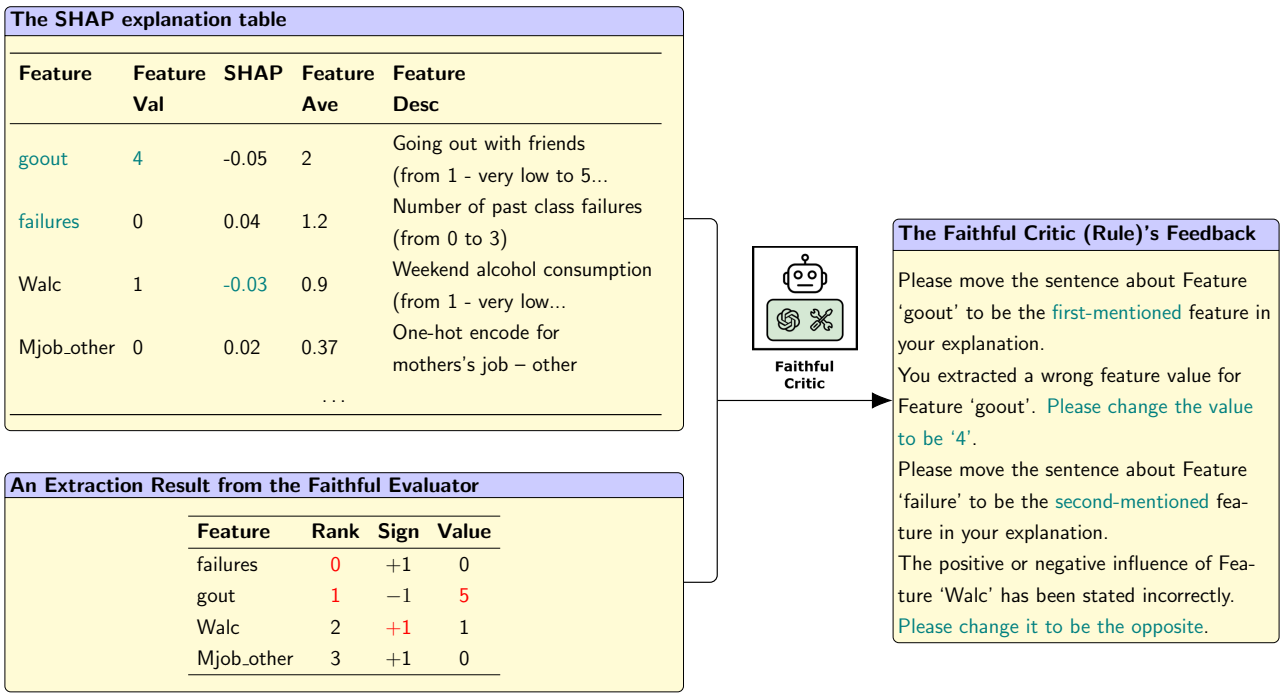}}
\caption{\textbf{Upper Left:} The instance's SHAP input table. \textbf{Below Left:} The extraction dictionary generated by the \textit{Faithful Evaluator}. All mistakes are successfully extracted and shown in red. \textbf{Right:} The output from the \textit{Faithful Critic}. It refers to the original SHAP table and outputs directional feedback, guiding the \textit{Narrator} on how to revise.}
\label{FC}
\end{figure}

\textbf{The Faithful Critic} The \textit{Faithful Critic} is designed to provide directional revision guidance based on the reported errors from the \textit{Faithful Evaluator}. Given the comparison result generated by the \textit{Faithful Evaluator}, the \textit{Faithful Critic} refers to the original SHAP table for the corresponding correct rank, sign, and value, and then provides clear and directional feedback. For example, when there is a rank problem identified in the narrative, the \textit{Faithful Critic} specifies how to relocate sentences to the correct place. It also guides the \textit{Narrator} to flip the influence from positive to negative (or vice versa) and to revise feature values to match those in the SHAP output when inconsistencies are detected. If no violation is detected, the \textit{Faithful Critic} returns the same message as the \textit{Faithful Evaluator}: ``After checking, the narrative is 100\% faithful to the SHAP table." An example is shown in Figure~\ref{FC}.

In principle, the formulation of the \textit{Faithful Critic}’s feedback does not require an LLM due to its rule-based nature. However, when multiple errors occur within a single narrative, the resulting feedback can become verbose and repetitive. Therefore, we introduce an optional LLM that functions solely as a summary agent, making the feedback into a more concise and coherent form, without losing any key information. We specially refer to this non-LLM variant as the \textit{Faithful Critic (Rule)}, and the original version as the \textit{Faithful Critic}. These two variants also enable a direct comparison of the performance benefits of incorporating an LLM.

\textbf{The Coherence Agent} The \textit{Coherence Agent} analyses the coherence issues existing in the narrative and provides corresponding revision instructions. Specifically, the \textit{Coherence Agent} evaluates whether the narrative flows smoothly, avoids abrupt transitions or disjoint statements, and is well structured overall. This is of interest because, beyond correctness that is represented by the faithfulness metric, the overall linguistic quality is also critical. Although LLMs are generally capable of producing fluent and coherent text, explicit guidance on improving coherence can make the revision process more systematic and direct.

\begin{figure}[htb]
\centering
\scalebox{0.55}{
\input{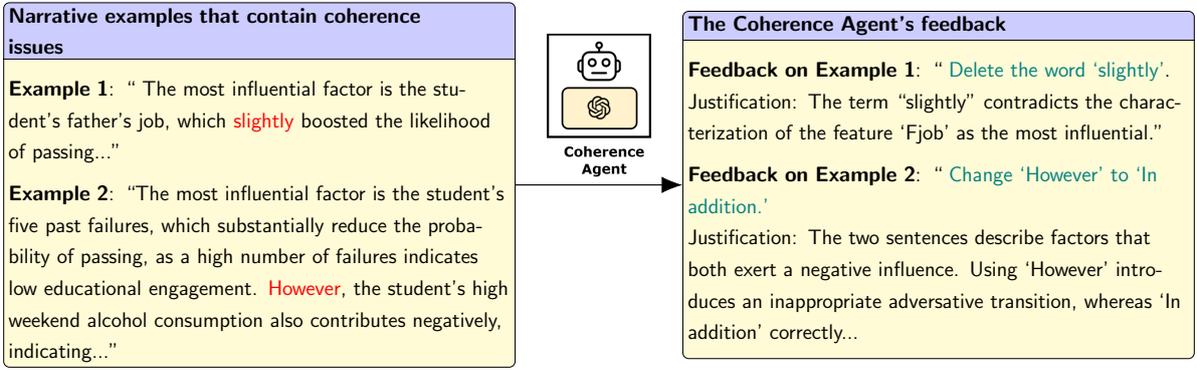}}
\caption{Two examples of the feedback from the \textit{Coherence Agent}. \textbf{Left:} Two narratives that contain coherence-related issues. Problems are highlighted in red. \textbf{Right:} The corresponding feedback from the \textit{Coherence Agent}.}
\label{CO}
\end{figure}

The \textit{Coherence Agent} prompt is intentionally simple and follows the same general structure as the other agents’ prompts. It begins with contextual information defining the agent’s role and task, followed by an explicit definition of \textit{coherence} adapted from~\cite{shen-etal-2023-large,Zemla2017-hw,electronics14132735}, the input narrative to be evaluated, and a predefined output structure. The \textit{Coherence Agent}’s output is regulated to follow four templates of revisions: ``Change \_\_\_ to \_\_\_”, ``Insert \_\_\_ before \_\_\_”, ``Delete \_\_\_”, ``Reorder \_\_\_ after \_\_\_”, following with justification below each modification suggestion.  If the \textit{Coherence Agent }doesn't find any coherence-related issues, it returns ``no coherence issues" with reasons. We show a few examples in Figure~\ref{CO}, and more analysis on different types of the \textit{Coherence Agent}'s feedback is in Section~\ref{coanalysis}.

\subsection{Agentic System Schematic}\label{architecture}

With all agents defined, their interactions are illustrated in Figure~\ref{schematic}. This figure depicts the most comprehensive system design, incorporating all four agents. We use pre-generated baseline narratives as round-0 inputs in this work to make sure a consistent starting point across all settings. These baseline narratives are fed directly into the \textit{Faithful Evaluator} and the \textit{Coherence Agent}, and the \textit{Narrator} then revises them based on round-0 feedback. Two different stopping criteria are employed, depending on the system design: when the \textit{Coherence Agent} is not included, the refinement process terminates either when 100\% faithfulness is achieved or when a predefined maximum number of rounds is reached; in contrast, when the \textit{Coherence Agent} is present, early stopping based on faithfulness is disabled, and all narratives must undergo the same turns until the maximum round is reached. This is particularly designed to examine how the \textit{Coherence Agent} modifies narratives that are already faithful but exhibit coherence issues.

We design multiple agentic systems with the attendance of different agents, with the intention to explore the impact from each agent and the performance for different configurations. The \textit{Narrator} and the \textit{Faithful Evaluator} are fixed components for every agentic system, with the \textit{Faithful Critic} and the \textit{Coherence Agent} being additional players. Since we have two \textit{Faithful Critic} variants, we in total have five designs:
    
\textbf{Basic Design:}  \textit{Narrator} +  \textit{Faithful Evaluator} 

\textbf{Critic Design:}  \textit{Narrator} +  \textit{Faithful Evaluator} +  \textit{Faithful Critic} 
    
\textbf{Critic-Rule Design:}  \textit{Narrator} +  \textit{Faithful Evaluator} +  \textit{Faithful Critic (Rule)}
    
\textbf{Coherent Design:}  \textit{Narrator} +  \textit{Faithful Evaluator} +  \textit{Faithful Critic} +  \textit{Coherence Agent}

\textbf{Coherent-Rule Design:}  \textit{Narrator} +  \textit{Faithful Evaluator} + \textit{Faithful Critic (Rule)} +  \textit{Coherence Agent}

\bigskip
\textbf{Basic Design}: When only the \textit{Faithful Evaluator} is included as the evaluator for faithfulness, its performance can be compared against the \textit{Critic Design} or the \textit{Critic-Rule Design} to validate the effectiveness of an extra faithfulness critic agent. This comparison mostly allows us to assess whether directional revision guidance is necessary for the \textit{Narrator} to efficiently correct unfaithful narratives. We expect that the refinement process in the \textit{Basic Design} is less efficient than in the \textit{Critic Design} and the \textit{Critic-Rule Design}.

\textbf{Critic Design and Critic-Rule Design}: The \textit{Critic Design} and the \textit{Critic-Rule Design} serve as two mirror groups. As stated above, they can both compare with the \textit{Basic Design}. Their comparison with each other allows us to evaluate whether incorporating an LLM into the \textit{Faithful Critic} is necessary.

\textbf{Coherent Design and Coherent-Rule Design}: The \textit{Coherent Design} and the \textit{Coherent-Rule Design} can compare with each other or with the \textit{Critic Design} and the \textit{Critic-Rule Design}, respectively. The comparison reflects how the \textit{Coherence Agent} impacts the faithfulness quality. We anticipate that the inclusion of the \textit{Coherence Agent} negatively affects faithfulness (Section~\ref{categorization}, C4), while it can also help identify subtle or minor textual issues that would otherwise remain unaddressed in systems without the \textit{Coherence Agent}.

\subsection{Evaluation Metrics}\label{evaluation}

\textbf{Faithfulness Evaluation} In this work, we mainly focus on the faithfulness metric to evaluate the narrative. We acknowledge that faithfulness can be measured in different ways, for example, a human-centered user study, LLM-as-a-judge based on a pre-defined rubric, or a more automated way that evaluates the correctness against the explainer's raw output~\cite{LLMXAINarrative}. This study, due to its automated nature, is more suited to the third approach. We hence adapted the method from~\cite{Howgoodismystory}, proposing \textit{Rank Accuracy (RA)}, \textit{Sign Accuracy (RA)}, and \textit{Value Accuracy (SA)}. These three metrics are simply evaluating the accuracy that $r_j$, $s_j$, and $v_j$ relative to the ground truth value $r^*_j, s^*_j, v^*_j$ respectively, where for example $r^*_j$ represents the actual rank of the extracted feature $f_j$.
(For XA is either RA/SA/VA):

\begin{equation}
XA = \frac{1}{M \cdot n} \sum_{i=1}^{M} \sum_{j=0}^{n-1} 
\begin{cases} 
1 & \text{if } x_{i,j} = \phi \\ 
\delta_{x_{i,j}, x_{i,j}^*} & \text{if } x_{i,j} \neq \phi 
\end{cases}
\label{xa}
\end{equation}

Where:
\begin{itemize}
    \item $M$ is the number of instances (narratives).
    \item $n$ is the pre-defined number of features per narrative. In the main experiments of this study, n=4.
    \item $x_{i,j}$ is the $(j+1)$-th feature of the $i$-th narrative, where $j \in \{0, \dots, n-1\}$.
\end{itemize}
\bigskip

Using this formula, the RA, SA, and VA are computed for each experimental round. The progression of these three metrics from round 0 to the final round illustrates the improvement in faithfulness achieved by the agent system. 

As mentioned before, we allow the \textit{\textit{Faithful Evaluator}} to decide whether to extract a feature value when a narrative does not explicitly specify it. In such cases, the value field in the extraction dictionary can possibly be \textit{nan} when the narrative provides only a qualitative description of the feature value (e.g., “many” or “some”) rather than an exact number. Due to this, a null value in the extraction dictionary is regarded as a valid and correct outcome and is added into the accuracy calculation. Additionally, the interpretation of rank accuracy (RA) differs from SA and VA. Once a single feature's rank is extracted incorrectly, it normally means at least one other feature's rank is swapped with it, which means one rank error is counted twice using Eq.(\ref{xa}). 

\textbf{Coherence Evaluation} As noted earlier, the plausibility (coherence) metric based on ``perplexity" proposed in~\cite{Howgoodismystory} proves less effective than expected. In a pilot experiment, we replicated this approach and similarly obtained uninformative results, with negligible and uninterpretable differences between baseline and refined narratives. We therefore excluded this ``perplexity" method from our evaluation. At the same time, although LLM-as-a-judge has emerged as a scalable and quantitative evaluation paradigm, it is highly sensitive to prompt design~\cite{LLMXAINarrative} and can yield inconsistent judgments~\cite{wang2025trustjudgeinconsistenciesllmasajudgealleviate}. As a result, we conduct a qualitative analysis to understand the impact of the \textit{Coherence Agent} on narrative quality, using representative examples and manual inspection.

\subsection{Experiments}

\textbf{Experiments Set-up} Our experimental framework uses five binary classification datasets: \textit{Student Performance Dataset} (exam pass or not), \textit{Fifa Man of the Match} (the winner of Man of the Match or not), \textit{German Credit Score }(bank credit good or bad), \textit{Diabetes Dataset} (get diabetes or not) and \textit{Stroke Prediction Dataset} (get a stroke or not). The first three datasets are from \cite{martens2024tellstorynarrativedrivenxai}, and the \textit{Diabetes}\footnote{https://www.kaggle.com/datasets/mathchi/diabetes-data-set}  and \textit{Stroke}\footnote{https://www.kaggle.com/datasets/fedesoriano/stroke-prediction-dataset}  datasets are from Kaggle. These datasets are chosen for their high feature interpretability, a reasonable number of features (8-40), and broad domain coverage. Following the methodology described by \cite{Howgoodismystory}, we randomly sampled 20 instances from each dataset's test set (10 per ground-truth class), yielding a total of 100 instances. We trained a Random Forest (RF) classifier as a prediction model, with the default scikit-learn hyperparameters. All LLMs are set to a temperature of $T=0$ to maximize reproducibility, though we acknowledge that their stochasticity can still lead to variation in their responses. For a prior test on DeepSeek-Chat-0324, we observe that most faithfulness issues can be resolved within 3 rounds or remain unresolved; therefore, we set the maximum number of iterations to 3 for all main experiments (we also have additional experiments where we set the maximum round to 10; see Section~\ref{addition}). Five LLMs are used: Claude Sonnet 4.5, Mistral Medium 3.1, GPT-5, LLama-3.3-70B, and DeepSeek-V3.2-Exp. These models are selected because they represent different geographic origins, open- or closed-source, and usage cost. Wherever possible, specific checkpoint versions were chosen to ensure reproducibility. We access Llama and DeepSeek via OpenRouter, while the remaining models are accessed through their official APIs. Details are provided in Appendix~\ref{LLMmodelselection}.

\textbf{Baseline Experiment} We conduct baseline experiments before the experiments on the agentic system. The baseline experiment follows a setting similar to~\cite{Howgoodismystory}, with a Generation Model to generate narratives and an Extraction Model to extract the information, as the \textit{Faithful Evaluator} does. The generation and extraction is one-time and no improvement process happens in this baseline experiment. We use DeepSeek-V3-0324 to generate 100 narratives across four runs and select the least-performing one as the baseline narratives for all subsequent experiments. Using baseline narratives ensures a consistent starting point across all experimental settings. Additionally, using the worst generation provides greater room for improvement in subsequent agentic experiments.

\textbf{Original Experiment} We first run experiments on all five agentic system designs once on Claude Sonnet 4.5, by which we compare their different improvement abilities and revision details (Section~\ref{originalclaude}). We then repeat this once for each of the remaining four LLMs, obtaining a complete set of results for all LLMs and agentic system designs (Section~\ref{originalall}). Once an LLM is selected, it is used consistently across all agents within the system.

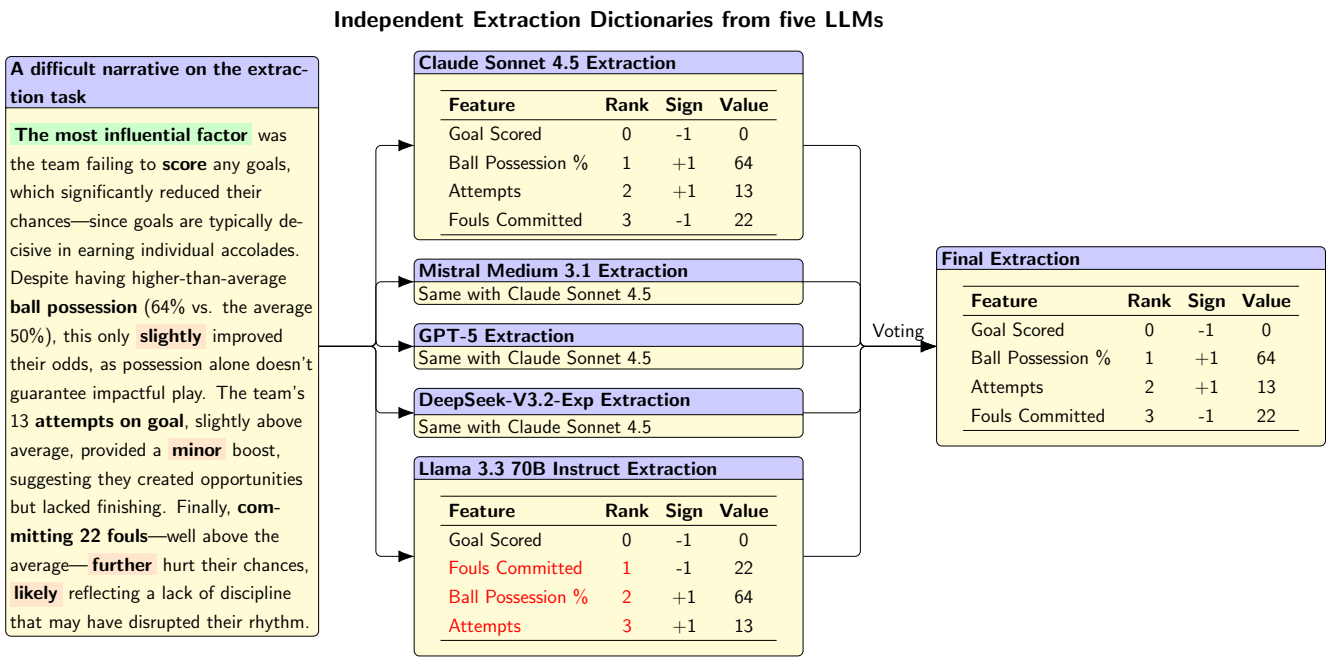
\begin{figure}[htb!]
\centering
\scalebox{0.50}{
\begin{tikzpicture}[node distance=2cm, font=\Large]

\tikzset{
    contentnode/.style n args={3}{
        draw,
        rounded corners,
        align=left,
        text width=#1,
        rectangle split,
        rectangle split parts=2,
        rectangle split part fill={blue!20, #2},
        rectangle split part align={center,left},
        #3
    }
}

\node[contentnode={8cm}{yellow!20}{}] (leftbox)
{
    \textbf{A difficult narrative on the extraction task}
    \nodepart{second}

    \colorbox{green!20}{\textbf{The most influential factor}} was the team failing to \textbf{score} any goals, which significantly reduced their chances—since goals are typically decisive in earning individual accolades. Despite having higher-than-average \textbf{ball possession} (64\% vs. the average 50\%), this only \colorbox{orange!20}{\textbf{slightly}} improved their odds, as possession alone doesn't guarantee impactful play. The team's 13 \textbf{attempts on goal}, slightly above average, provided a \colorbox{orange!20}{\textbf{minor}} boost, suggesting they created opportunities but lacked finishing. Finally, \textbf{committing 22 fouls}—well above the average—\colorbox{orange!20}{\textbf{further}} hurt their chances, \colorbox{orange!20}{\textbf{likely}} reflecting a lack of discipline that may have disrupted their rhythm.

    \vspace{0.5em}
};

\node[contentnode={10cm}{yellow!20}{}, right=2.5cm of leftbox.east] (em3)
{
    \textbf{GPT-5 Extraction}
    \nodepart{second}
    Same with Claude Sonnet 4.5
};

\node[contentnode={10cm}{yellow!20}{}, above=0.5cm of em3] (em2)
{
    \textbf{Mistral Medium 3.1 Extraction}
    \nodepart{second}
    Same with Claude Sonnet 4.5
};

\node[contentnode={10cm}{yellow!20}{}, above=0.5cm of em2] (em1)
{
    \textbf{Claude Sonnet 4.5 Extraction}
    \nodepart{second}
        \begin{center}
        \begin{tabular}{lccc}
        \hline
        \textbf{Feature} & \textbf{Rank} & \textbf{Sign} & \textbf{Value} \\
        \hline
        Goal Scored & 0 & -1 & 0 \\
        Ball Possession \%     & 1 & +1 & 64 \\
        Attempts      & 2 &+1 & 13 \\
        Fouls Committed     & 3 & -1 & 22 \\
        \hline
        \end{tabular}
        \end{center}
};

    \node[contentnode={10cm}{yellow!20}{}, below=0.5cm of em3] (em4)
{
    \textbf{DeepSeek-V3.2-Exp Extraction}
    \nodepart{second}
    Same with Claude Sonnet 4.5
};

    \node[contentnode={10cm}{yellow!20}{}, below=0.5cm of em4] (em5)
{
    \textbf{Llama 3.3 70B Instruct Extraction}
    \nodepart{second}
        \begin{center}
        \begin{tabular}{lccc}
        \hline
        \textbf{Feature} & \textbf{Rank} & \textbf{Sign} & \textbf{Value} \\
        \hline
        Goal Scored & 0 & -1 & 0 \\
        \textcolor{red}{Fouls Committed}     & \textcolor{red}{1} & -1 & 22 \\
        \textcolor{red}{Ball Possession \%}     & \textcolor{red}{2} & +1 & 64 \\
        \textcolor{red}{Attempts}      & \textcolor{red}{3} &+1 & 13 \\
        \hline
        \end{tabular}
        \end{center}
\vspace{0.5em}

};

\node[contentnode={10cm}{yellow!20}{}, right=3.5cm of em3.east, anchor=west] (feedback0)
{
    \textbf{Final Extraction}
    \nodepart{second}
        \begin{center}
        \begin{tabular}{lccc}
        \hline
        \textbf{Feature} & \textbf{Rank} & \textbf{Sign} & \textbf{Value} \\
        \hline
        Goal Scored & 0 & -1 & 0 \\
        Ball Possession \%     & 1 & +1 & 64 \\
        Attempts      & 2 &+1 & 13 \\
        Fouls Committed     & 3 & -1 & 22 \\
        \hline
        \end{tabular}
        \end{center}
     \vspace{0.5em}
};

\node[above=0.4cm of em1, font=\LARGE\bfseries] {Independent Extraction Dictionaries from five LLMs};

\foreach \i in {1,2,3,4,5} {
    \draw[-{Latex[length=4mm, width=3mm]}, rounded corners] (leftbox.east) -- ++(1.5,0) |- (em\i.west);
    
    \draw[-{Latex[length=4mm, width=3mm]}, rounded corners] (em\i.east) -- ++(1.5,0) |- (feedback0.west);
};

\node at ($(em3.east) + (2.5,0)$) [above=2pt, inner sep=2pt, font=\Large] {Voting};

\end{tikzpicture}}
\caption{\textbf{Left:} A narrative that is linguistically ambiguous on the importance of features. This is a difficult example for the \textit{Faithful Evaluator} to correctly extract all rank information. \textbf{Middle:} Five LLM-based \textit{Faithful Evaluator} extracts the information independently and a disagreement happens. For example, Llama 3.3 70B Instruct makes mistakes on the extraction dictionary (mark in red), whereas other LLMs do not. \textbf{Right:} With the ensemble strategy (majority voting), the extraction results from five LLMs are combined to make a final decision and mitigate extraction mistakes. }
\label{ESM}
\end{figure}

\textbf{Ensemble Experiment} According to the result from original experiments, we observe that the \textit{Faithful Evaluator} makes \textit{extraction mistakes} across all LLMs, though some LLMs perform better than others. An \textit{extraction mistake} refers to cases where the \textit{Faithful Evaluator} extracts feature rank, sign, or value information that does not correctly reflect the content of the narrative, as shown in Figure~\ref{ESM}.  These \textit{extraction mistakes} can technically be of two types: \textit{false positives}, where an unfaithful narrative is judged to be faithful, and \textit{false negatives}, where a faithful narrative is incorrectly reported with errors. False negative \textit{extraction mistakes} are observed in our original experiments, leading to lower reported accuracy than in reality\footnote{We note that \textit{false positives} are more subtle than \textit{false negatives}, as unreported narratives are far more numerous, making manual verification impractical. However,\textit{ false positives} occur not as often as \textit{false negatives}. This is supported by the validation experiments from~\cite{Howgoodismystory}, where the Extraction LLM (GPT-4o-based) produces \textit{false negatives} by incorrectly flagging faithful narratives as faulty, while producing zero \textit{false positives}. For this reason, we intentionally exclude \textit{false positive} cases from this study. }. In this study, we use the term \textit{extraction mistake} to specifically refer to this mistake in the \textit{false negative} kind.

With the intention to solve this issue, we propose an ensemble strategy, in which the sole \textit{Faithful Evaluator} is replaced by a group of \textit{Faithful Evaluator} voters, each based on one of the five LLMs. All five \textit{Faithful Evaluators} independently extract the narrative's information, and their extraction dictionaries are aggregated by majority voting to produce a final dictionary. In detail, all extraction dictionaries are collected, and only the most common feature set (keys) is retained in the final dictionary. Once an LLM-based \textit{Faithful Evaluator}'s extraction contains different feature names or other strange situations, this extraction is excluded from voting. When there is a tie in the voting on either the ``rank", ``sign", or ``value" field, the final extraction adopts the value that appears first. The other configuration of the ensemble experiment is no different than the original one. All five system designs are executed (Section~\ref{ensemble}), as in the original experiment. We expect the ensemble experiments to consistently outperform the original experiments because the ensemble approach reduces extraction errors that would otherwise lower the reported accuracy. 

\textbf{Additional Experiments on Key Concepts} We further investigate the influence of several key concepts on the effectiveness of the agentic system. Those key concepts are listed in Table~\ref{additionaltable}.

\begin{table*}[h!]
\centering
\caption{The key concepts of the additional experiments}
\label{additionaltable}
\footnotesize 
\begin{tabular}{l p{5cm} p{3cm} p{4cm}}
\toprule
\textbf{Concepts} & \textbf{Reason} & \textbf{Original Setting}  & \textbf{New Setting} \\
\midrule

\textbf{Iteration Round} & How does the number of iteration rounds impact the result? & Round=2 & Round=10\\
\midrule
\textbf{The Number of Features} & How does the number of features included in the narrative impact the result? & Feature=4 & Feature=8 \\
\midrule
\textbf{No-Baseline Setting} &What if we don't use baseline narratives? How does this impact the result? & Baseline narratives from Deepseek-chat-0324& No baseline narratives\\
\midrule
\textbf{Different Baseline Narratives} &How do the best and the worst baseline narratives impact the result? & Baseline narratives from Deepseek-chat-0324 & \makecell[l]{1. Best Baseline narratives\\from DeepSeek-V3.2-Exp\\2. Worst Baseline narratives\\from Llama 3.3 70B Instruct}\\
\bottomrule
\end{tabular}
\end{table*}

In order to test those key concepts, we select a single "best" agentic system design (the \textit{Critic-Rule Design}). The rationale for choosing this design and the results are detailed in Section~\ref{addition}. The set of baseline narratives containing eight features is generated in the same manner as in the previous experiments, using DeepSeek-chat-0324. The best and worst baseline narratives are determined based on the results obtained under the no-baseline setting. For each of the four key concepts, only the corresponding condition is modified, while all other configurations remain identical to those of the original \textit{Critic-Rule Design.}

\section{Results and Analysis}
\subsection{Faithfulness Results for Claude Sonnet 4.5}\label{originalclaude}
\subsubsection{The Improvement of Rank, Sign and Value Accuracy}\label{main}

\begin{figure}[!b]
    \centering
\includegraphics[width=1.0\textwidth]{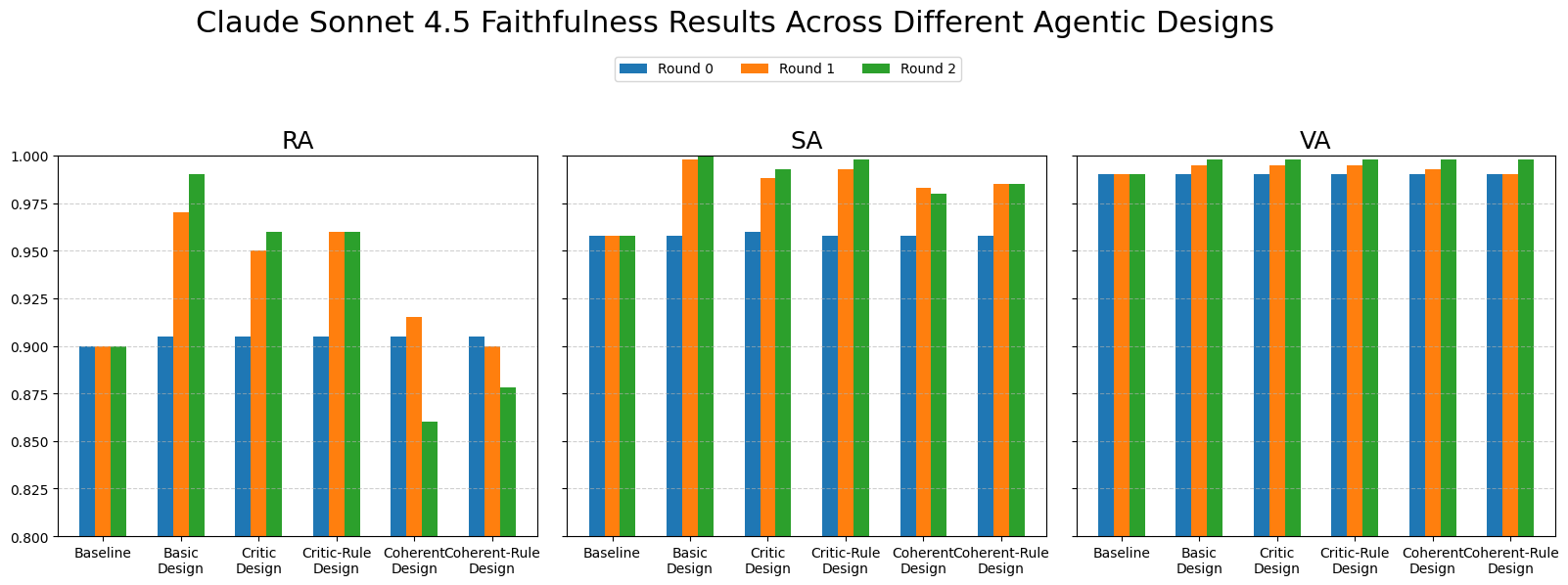}
    \caption{This bar chart represents the three accuracies (\textit{RA}: rank accuracy,\textit{ SA}: sign accuracy, and \textit{VA}: value accuracy) for different agentic system designs across the three rounds using Claude Sonnet 4.5.}
    \label{claudebar}
\end{figure}

Table~\ref{claudetable} and Figure~\ref{claudebar} present the faithfulness results for Claude Sonnet 4.5 across all five agentic system designs.  The \textit{Basic Design}, the \textit{Critic Design}, and the \textit{Critic-Rule Design} demonstrate significant improvements over the three refinement rounds, whereas the \textit{Coherent Design} and the \textit{Coherent-Rule Design} show a drop in rank accuracy on four datasets. The number of unfaithful narratives decreases for all five designs, with varying degrees of reduction.

\begin{table}[h!]
    \scriptsize
    \centering
    \caption{
Faithfulness results for different agentic system designs based on Claude Sonnet 4.5. Arrows indicate the progression of the rank, sign, and value accuracy from Round 0 to Round 2. The last column shows the progression of the number of unfaithful narratives over rounds.}
    \begin{tabular}{l l c c c c}
        \toprule
        \multirow{2}{*}{\makecell[c]{\textbf{Claude Sonnet 4.5}\\ \textbf{based Designs}}} & \multirow{2}{*}{\textbf{Datasets}} & \multicolumn{3}{c}{\textbf{Faithfulness}} 
        & \multirow{2}{*}{\makecell[c]{\textbf{No. Unf.} \\ \textbf{Narr.} }}\\
        \cmidrule(r){3-5}  
        &  & \textbf{RA} & \textbf{SA} & \textbf{VA} &
        \\
        \midrule
        \multirow{6}{*}{\textbf{\makecell[l]{Baseline}}} &  
        \textbf{AVG.} &
        \textbf{(0.900}) & 
        \textbf{0.958} & 
        \textbf{0.990} &
        \textbf{(31)} \\ 
        \textbf{} & 
        diabetes &
        0.838 & 0.988 & 1.000 & 5 \\
        \textbf{} & 
        stroke &
        (0.925) & 0.938 & 0.975 & (8) \\
        \textbf{} & 
        student &
        0.925 & 0.913 & 1.000 & 8 \\
        \textbf{} & 
        fifa &
        0.875 & 0.963 & 1.000 & 7 \\
        \textbf{} & 
        credit &
        0.938 & 0.988 & 0.975 & 3 \\
        \hline
        \multirow{6}{*}{\textbf{\makecell[l]{Basic Design}}} & \textbf{AVG.} &        
        \textbf{ 0.905 $\rightarrow$ 0.970 $\rightarrow$ 0.990} & 
        \textbf{ 0.958 $\rightarrow$ 0.998 $\rightarrow$ 1.000} & 
        \textbf{ 0.990 $\rightarrow$ 0.995 $\rightarrow$ 0.998} & 
        \textbf{ 30 $\rightarrow$ 6 $\rightarrow$ 3 } \\
        \textbf{} & 
        diabetes &
        0.838 $\rightarrow$ 0.950 $\rightarrow$ 0.975 & 
        0.988 $\rightarrow$ 1.000 $\rightarrow$ 1.000 & 
        1.000 $\rightarrow$ 1.000 $\rightarrow$ 1.000 &
        5 $\rightarrow$ 2 $\rightarrow$ 1 \\
        \textbf{} & 
        stroke &
        0.950 $\rightarrow$ 0.988 $\rightarrow$ 1.000 & 
        0.938 $\rightarrow$ 0.988 $\rightarrow$ 1.000 & 
        0.975 $\rightarrow$ 1.000 $\rightarrow$ 1.000 &
        7 $\rightarrow$ 1 $\rightarrow$ 0 \\
        \textbf{} & 
        student &
        0.925 $\rightarrow$ 0.975 $\rightarrow$ 0.975 & 
        0.913 $\rightarrow$ 1.000 $\rightarrow$ 1.000 & 
        1.000 $\rightarrow$ 1.000 $\rightarrow$ 1.000 &
        8 $\rightarrow$ 1 $\rightarrow$ 1 \\
        \textbf{} & 
        fifa &
        0.875 $\rightarrow$ 0.963 $\rightarrow$ 1.000 & 
        0.963 $\rightarrow$ 1.000 $\rightarrow$ 1.000 & 
        1.000 $\rightarrow$ 1.000 $\rightarrow$ 1.000 &
        7 $\rightarrow$ 1 $\rightarrow$ 0 \\
        \textbf{} & 
        credit &
        0.938 $\rightarrow$ 0.975 $\rightarrow$ 1.000 & 
        0.988 $\rightarrow$ 1.000 $\rightarrow$ 1.000 & 
        0.975 $\rightarrow$ 0.975 $\rightarrow$ 0.988 &
        3 $\rightarrow$ 1 $\rightarrow$ 1 \\
        \hline
        \multirow{6}{*}{\textbf{\makecell[l]{Critic Design}}} & \textbf{AVG.} &        
        \textbf{ 0.905 $\rightarrow$ 0.950 $\rightarrow$ 0.960} & 
        \textbf{ (0.960) $\rightarrow$ 0.988 $\rightarrow$ 0.993} & 
        \textbf{ 0.990 $\rightarrow$ 0.995 $\rightarrow$ 0.998} & 
        \textbf{ 30 $\rightarrow$ 13 $\rightarrow$ 9 } \\
        \textbf{} & 
        diabetes &
        0.838 $\rightarrow$ 0.950 $\rightarrow$ 0.950 & 
        (1.000) $\rightarrow$ 0.988 $\rightarrow$ 1.000 & 
        1.000 $\rightarrow$ 1.000 $\rightarrow$ 1.000 &
        5 $\rightarrow$ 2 $\rightarrow$ 2 \\
        \textbf{} & 
        stroke &
        0.950 $\rightarrow$ 0.975 $\rightarrow$ 0.975 & 
        0.938 $\rightarrow$ 0.988 $\rightarrow$ 0.988 & 
        0.975 $\rightarrow$ 0.988 $\rightarrow$ 0.988 &
        7 $\rightarrow$ 2 $\rightarrow$ 2 \\
        \textbf{} & 
        student &
        0.925 $\rightarrow$ 0.975 $\rightarrow$ 0.975 & 
        0.913 $\rightarrow$ 0.975 $\rightarrow$ 0.975 & 
        1.000 $\rightarrow$ 1.000 $\rightarrow$ 1.000 &
        8 $\rightarrow$ 2 $\rightarrow$ 2 \\
        \textbf{} & 
        fifa &
        0.875 $\rightarrow$ 0.875 $\rightarrow$ 0.900 & 
        0.963 $\rightarrow$ 0.988 $\rightarrow$ 1.000 & 
        1.000 $\rightarrow$ 1.000 $\rightarrow$ 1.000 &
        7 $\rightarrow$ 5 $\rightarrow$ 3 \\
        \textbf{} & 
        credit &
        0.938 $\rightarrow$ 0.975 $\rightarrow$ 1.000 & 
        0.988 $\rightarrow$ 1.000 $\rightarrow$ 1.000 & 
        0.975 $\rightarrow$ 0.988 $\rightarrow$ 1.000 &
        3 $\rightarrow$ 2 $\rightarrow$ 0 \\
        \hline
        \multirow{6}{*}{\textbf{\makecell[l]{Critic-Rule Design}}} & \textbf{AVG.} &        
        \textbf{ 0.905 $\rightarrow$ 0.960 $\rightarrow$ 0.960} & 
        \textbf{ 0.958 $\rightarrow$ 0.993 $\rightarrow$ 0.998} & 
        \textbf{ 0.990 $\rightarrow$ 0.995 $\rightarrow$ 0.998} & 
        \textbf{ 30 $\rightarrow$ 10 $\rightarrow$ 8 } \\
        \textbf{} & 
        diabetes &
        0.838 $\rightarrow$ 0.975 $\rightarrow$ 0.975 & 
        0.988 $\rightarrow$ 1.000 $\rightarrow$ 1.000 & 
        1.000 $\rightarrow$ 1.000 $\rightarrow$ 1.000 &
        5 $\rightarrow$ 1 $\rightarrow$ 1 \\
        \textbf{} & 
        stroke &
        0.950 $\rightarrow$ 0.975 $\rightarrow$ 0.975 & 
        0.938 $\rightarrow$ 0.988 $\rightarrow$ 0.988 & 
        0.975 $\rightarrow$ 1.000 $\rightarrow$ 0.988 &
        7 $\rightarrow$ 2 $\rightarrow$ 2 \\
        \textbf{} & 
        student &
        0.925 $\rightarrow$ 0.975 $\rightarrow$ 0.975 & 
        0.913 $\rightarrow$ 0.975 $\rightarrow$ 1.000 & 
        1.000 $\rightarrow$ 1.000 $\rightarrow$ 1.000 &
        8 $\rightarrow$ 2 $\rightarrow$ 1 \\
        \textbf{} & 
        fifa &
        0.875 $\rightarrow$ 0.900 $\rightarrow$ 0.900 & 
        0.963 $\rightarrow$ 1.000 $\rightarrow$ 1.000 & 
        1.000 $\rightarrow$ 1.000 $\rightarrow$ 1.000 &
        7 $\rightarrow$ 3 $\rightarrow$ 3 \\
        \textbf{} & 
        credit &
        0.938 $\rightarrow$ 0.975 $\rightarrow$ 0.975 & 
        0.988 $\rightarrow$ 1.000 $\rightarrow$ 1.000 & 
        0.975 $\rightarrow$ 0.975 $\rightarrow$ 1.000 &
        3 $\rightarrow$ 2 $\rightarrow$ 1 \\
        \hline
        \multirow{6}{*}{\textbf{\makecell[l]{Coherent Design}}} & \textbf{AVG.} &        
        \textbf{ 0.905 $\rightarrow$ 0.915 $\rightarrow$ 0.860} & 
        \textbf{ 0.958 $\rightarrow$ 0.983 $\rightarrow$ 0.980} & 
        \textbf{ 0.990 $\rightarrow$ 0.993 $\rightarrow$ 0.998} & 
        \textbf{ 30 $\rightarrow$ 20 $\rightarrow$ 28} \\
        \textbf{} & 
        diabetes &
        0.838 $\rightarrow$ 0.888 $\rightarrow$ 0.875 & 
        0.988 $\rightarrow$ 0.988 $\rightarrow$ 1.000 & 
        1.000 $\rightarrow$ 1.000 $\rightarrow$ 1.000 &
        5 $\rightarrow$ 4 $\rightarrow$ 4 \\
        \textbf{} & 
        stroke &
        0.950 $\rightarrow$ 0.938 $\rightarrow$ 0.838 & 
        0.938 $\rightarrow$ 0.988 $\rightarrow$ 0.963 & 
        0.975 $\rightarrow$ 0.988 $\rightarrow$ 0.988 &
        7 $\rightarrow$ 3 $\rightarrow$ 9 \\
        \textbf{} & 
        student &
        0.925 $\rightarrow$ 0.888 $\rightarrow$ 0.900 & 
        0.913 $\rightarrow$ 0.963 $\rightarrow$ 0.963 & 
        1.000 $\rightarrow$ 0.988 $\rightarrow$ 1.000 &
        8 $\rightarrow$ 5 $\rightarrow$ 5 \\
        \textbf{} & 
        fifa &
        0.875 $\rightarrow$ 0.925 $\rightarrow$ 0.750 & 
        0.963 $\rightarrow$ 0.975 $\rightarrow$ 0.975 & 
        1.000 $\rightarrow$ 1.000 $\rightarrow$ 1.000 &
        7 $\rightarrow$ 4 $\rightarrow$ 8 \\
        \textbf{} & 
        credit &
        0.938 $\rightarrow$ 0.938 $\rightarrow$ 0.938 & 
        0.988 $\rightarrow$ 1.000 $\rightarrow$ 1.000 & 
        0.975 $\rightarrow$ 0.988 $\rightarrow$ 1.000 &
        3 $\rightarrow$ 4 $\rightarrow$ 2 \\
        \hline
        \multirow{6}{*}{\textbf{\makecell[l]{Coherent-Rule Design}}} & \textbf{AVG.} &        
        \textbf{ 0.905 $\rightarrow$ 0.900 $\rightarrow$ 0.878} & 
        \textbf{ 0.958 $\rightarrow$ 0.985 $\rightarrow$ 0.985} & 
        \textbf{ 0.990 $\rightarrow$ 0.990 $\rightarrow$ 0.998} & 
        \textbf{ 30 $\rightarrow$ 22 $\rightarrow$ 25 } \\
        \textbf{} & 
        diabetes &
        0.838 $\rightarrow$ 0.838 $\rightarrow$ 0.925 & 
        0.988 $\rightarrow$ 0.988 $\rightarrow$ 0.988 & 
        1.000 $\rightarrow$ 1.000 $\rightarrow$ 1.000 &
        5 $\rightarrow$ 5 $\rightarrow$ 4 \\
        \textbf{} & 
        stroke &
        0.950 $\rightarrow$ 0.938 $\rightarrow$ 0.863 & 
        0.938 $\rightarrow$ 0.988 $\rightarrow$ 0.963 & 
        0.975 $\rightarrow$ 0.975 $\rightarrow$ 0.988 &
        7 $\rightarrow$ 4 $\rightarrow$ 7 \\
        \textbf{} & 
        student &
        0.925 $\rightarrow$ 0.888 $\rightarrow$ 0.875 & 
        0.913 $\rightarrow$ 0.963 $\rightarrow$ 1.000 & 
        1.000 $\rightarrow$ 0.988 $\rightarrow$ 1.000 &
        8 $\rightarrow$ 5 $\rightarrow$ 4 \\
        \textbf{} & 
        fifa &
        0.875 $\rightarrow$ 0.900 $\rightarrow$ 0.800 & 
        0.963 $\rightarrow$ 0.988 $\rightarrow$ 0.975 & 
        1.000 $\rightarrow$ 1.000 $\rightarrow$ 1.000 &
        7 $\rightarrow$ 4 $\rightarrow$ 7 \\
        \textbf{} & 
        credit &
        0.938 $\rightarrow$ 0.938 $\rightarrow$ 0.925 & 
        0.988 $\rightarrow$ 1.000 $\rightarrow$ 1.000 & 
        0.975 $\rightarrow$ 0.988 $\rightarrow$ 1.000 &
        3 $\rightarrow$ 4 $\rightarrow$ 3 \\
        \hline
    \end{tabular}
    
    \footnotesize{Note: A total of 100 baseline narratives from five datasets are used to make the starting point consistent across experiments. However, even when $T$=0, due to the LLM uncertainty, the \textit{Faithful Evaluator} may occasionally extract different results from the same narrative and hence leads to different accuracy and/or different number of unfaithful narratives in round 0. These differences are highlighted in brackets. }
\label{claudetable}
\end{table}

At the beginning, the average rank, sign, and value accuracies from the one hundred baseline narratives are 0.900, 0.958, and 0.990, respectively, with 31 narratives containing faithfulness-related issues. The \textit{Credit} dataset has the strongest baseline narratives, only 3 out of 20 containing faithfulness issues. Across all five designs, the \textit{Basic Design} appears to be the best agentic system for Claude Sonnet 4.5. By round 2, its average rank, sign, and value accuracies are the highest compared to others, reaching 0.990, 1.000, and 0.998, respectively. The \textit{Basic Design} is also the most efficient system. For example, the sign accuracy for four datasets has already reached 100\% in round 1. This system corrects all 40 narratives from the \textit{Stroke} and \textit{Fifa} datasets by the final round; only three narratives don't get fixed in the end. The \textit{Critic (Rule) Design} and the \textit{Critic Design} are both effective in improving the narratives, although not as good as the \textit{Basic Design}. The \textit{Critic (Rule) Design} has a minor advantage over the \textit{Critic Design}. The \textit{Coherence Design} and the \textit{Coherence-Rule Design} perform the worst in the faithfulness evaluation: they degrade rank accuracy after two rounds of refinement and leave a large number of unfaithful narratives (28 and 25, compared with the initial 30). This is expected because the \textit{Coherence Agent} is designed to improve the linguistic coherence quality, which may hinder the faithfulness quality. This is elaborated in the next section, under the problem category 4: \textit{Misleading Coherence Instruction}.

\subsubsection{Problem Categorization of Unfaithful Narratives}\label{categorization}

\begin{table}[b!]
\footnotesize
\centering
\caption{
Problem Categorization of Unfaithful Narratives}
\renewcommand{\arraystretch}{1.2}
    \begin{tabular}{p{5.0cm} p{11cm}}
    \hline
    \textbf{Category} & \textbf{Conditions} \\
    \midrule
    \textbf{\makecell[l]{C1: Confusing Faithfulness \\ Instruction}} & 
    \textbf{\textit{C1.1 }} Since the \textit{\textit{Faithful Evaluator}} feedback contains non-directional instruction, the \textit{Narrator} does not clearly know how to revise accordingly. This occurs because the \textit{\textit{Faithful Evaluator}} only reveals the error type by how it is designed, without mentioning where the \textit{Narrator} should change to. \\
    & \textbf{\textit{C1.2 (edge case)}} The \textit{\textit{Faithful Critic}} feedback contains information on reordering the mentioning of features while it already satisfies. This occurs when a narrative mentions all features following the order in the SHAP, while containing inappropriate descriptive phrases that violate the importance of that feature in the SHAP table (Figure~\ref{C1.2}). \\
        \midrule
    \textbf{C2: Extraction Mistake} & 
    The \textit{\textit{Faithful Evaluator}} reports non-existing faithfulness issues when a narrative is (arguably) faithful. 
    \\
        \midrule
    \textbf{\makecell[l]{C3: Counterintuitive Feature \\Influence}} &
    When the given SHAP input contains counterintuitive feature effects, the \textit{Narrator} is inclined to disregard and generate a narrative following common sense. \\
        \midrule
    \textbf{\makecell[l]{C4: Misleading Coherence \\Instruction}} &
    The \textit{\textit{Coherence Agent}} gives feedback with the intention to improve the narrative's coherence, but it inadvertently encourages revisions that violate the SHAP input. \\
        \midrule
    \textbf{C5: Generation Misalignment} & 
    The \textit{Narrator} fails to generate a faithful narrative with an identifiable cause from the four above. This is therefore attributed to one-time generation errors that LLMs may produce. \\
    \bottomrule
    \label{categorizationtable}
    \end{tabular}
\end{table}

To examine why faithfulness issues persist after three rounds for each design, we review all unfaithful narratives in round 2 and classify them into five categories according to the underlying reason (Table~\ref{categorizationtable}). We also categorize the 31 unfaithful baseline narratives using the same principles. 

\textbf{Category 1. Confusing Faithfulness Instruction}

The \textit{Narrator} doesn't successfully fix a narrative because it doesn't receive clear instructions. This situation occurs in two situations.

\textit{\textbf{C1.1 }} happens in the the \textit{Basic Design} system, where the \textit{\textit{Faithful Critic}} isn't present. Because the \textit{\textit{Faithful Evaluator}} only reports the existing errors, \textit{Narrator} does not directly get the information about how to revise the narrative accordingly. An example from \textit{Credit} dataset (index 10): the \textit{\textit{Faithful Evaluator}}'s feedback includes this sentence: ``Feature savings contains (an) errors in [`value'] value. " Since the \textit{\textit{Faithful Evaluator}} doesn't provide clear information on what value it should change to, \textit{Narrator} doesn't solve this problem after three rounds of trying. 

\textbf{\textit{C1.2 }} happens, on the contrary, when the \textit{\textit{Faithful Critic}} is present. Most of the time, since the \textit{\textit{Faithful Critic}} gives clear instructions on how to fix, the narrative should get fixed quickly. However, a rare case is when the \textit{Faithful Critic} feedback contains instructions that a narrative already satisfies, leading to ineffective revisions.

\begin{figure}[H]
\centering
\scalebox{0.50}{
\input{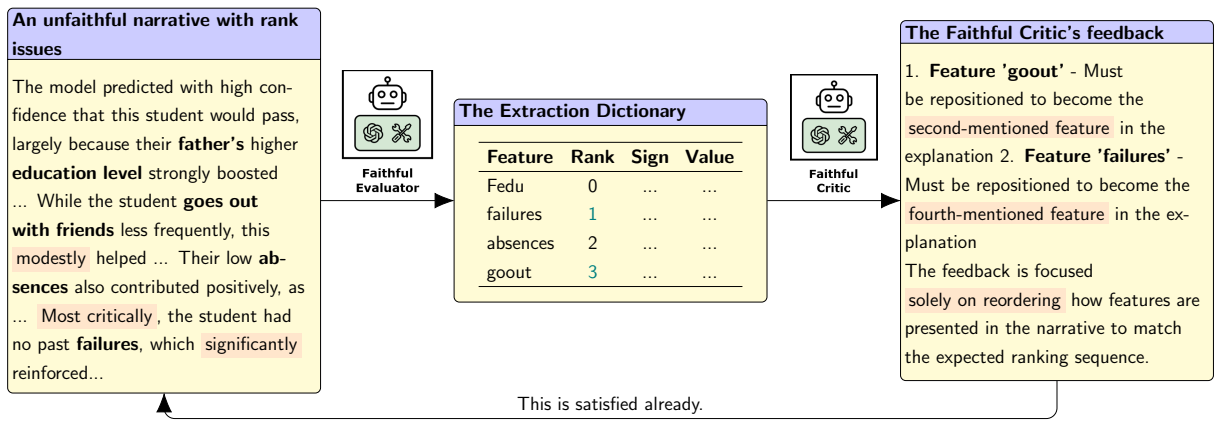}}
\caption{\textbf{Left:} This is an narrative example from the \textit{Student} dataset, index 14, from the Claude Sonnet 4.5-based \textit{Critic Design}. This narrative mentions four features following the feature rank in the SHAP input. However, this narrative is unfaithful because the inappropriate use of descriptive terms (highlighted in orange boxes) implies that the feature importance of ``failures" is higher than that of ``goout". This is an edge case in which the feature-mention order does not match the feature-importance order. \textbf{Middle:} The \textit{Faithful Evaluator} extracts the rank of ``failures" to be the second most important, and the feature ``goout" to be the least important. It does so because it does not mechanically determine the rank based on the order in which features are mentioned; instead, it infers the ranking through a holistic linguistic understanding of the narrative. \textbf{Right:} The \textit{Faithful Critic} instructs repositioning the two sentences. However, this is already satisfied in this narrative at the beginning.}
\label{C1.2}
\end{figure}

Consider an example from the \textit{Critic Design} (\textit{Student} dataset, index 14, Figure~\ref{C1.2}). In this case, the narrative mentions four features one by one, following their order in the SHAP table (rank from 0 to 3, representing the most important and the fourth most important feature). However, the sentence describing the fourth-mentioned and least important feature (“failures”)  begins with the phrase “most critically.” As a result, the \textit{Faithful Evaluator} extracts the rank of the feature ``failure" to be 1, representing the second most important feature by its overall understanding of the narrative, while this feature's actual rank is 3 in the SHAP table. Simultaneously, the \textit{Faithful Evaluator} assigns the feature ``goout” a rank of 3, while its actual rank in the SHAP table is 1. Up to this point, the narrative indeed contains a rank problem caused by the inappropriate term ``most critically", and the \textit{Faithful Evaluator} does not make any mistake. Afterwards, the \textit{Faithful Critic} detects this mismatch between the extracted ranks and the SHAP rankings and provides the feedback that requires the repositioning of the related sentences. This is where the issue arises, because the feedback from the \textit{Faithful Critic} suggests changing the feature-mention order that is already satisfied; even worse, it adds an additional constraint instructing the \textit{Narrator} not to modify anything other than the feature order (``...solely on reordering...") for some unknown reason. As a result, the \textit{Narrator} doesn't change a word during the iteration, and hence it fails to achieve 100\% faithful in round 2.

The root cause of category \textit{C1.2} is poorly generated baseline narratives that contain inappropriate descriptive terms that clearly contradict the feature importance indicated by SHAP. Another contributing factor lies in the design of the \textit{\textit{Faithful Critic}}. In its rank-correction component, it mechanically detects the rank differences and recommends relocating the feature accordingly. In other words, the \textit{Faithful Critic} is unable to capture the information from the inappropriate descriptive phrases; it merely produces feedback on feature relocation by its design. The \textit{\textit{Faithful Critic}} is designed in this way because of the intuitive assumption that a narrative mentions the feature from the most important to the least important, which generally holds in practice. Fortunately, the \textit{\textit{Faithful Critic}} is effective in most cases, and this problem occurs just twice among the 100 baseline narratives generated by Deepseek-chat-0324. Therefore, we retain the way we design the \textit{Faithful Critic} as is.

\textbf{Category 2. Extraction Mistake}

This situation occurs when {the \textit{\textit{Faithful Evaluator}} makes mistakes on extracting information from a (arguably) faithful narrative, reporting non-existing faithfulness issues. The first possibility arises from the stochastic behavior of the LLM. The second reason is that the narrative contains vague expressions that make it a genuinely difficult task for the \textit{\textit{Faithful Evaluator}}: for instance, not all narratives explicitly distinguish the importance of a feature using clear descriptive terms, thereby causing ambiguity and subjectivity in understanding and the \textit{Faithful Evaluator} find it hard to extract the correct feature rank. This is why we describe such cases as ``arguably" faithful: the narrative is debatably faithful, but its wording leaves room for ambiguity. An example is when adverbs or adjectives such as “slightly”, ``marginally", ``minor", and “small” are used to describe the importance of features, which show little differences among them (an example in Figure~\ref{EM}). In this case, the judgment of feature ranking is expected to be uncertain and debatable because subtle linguistic nuances can lead to inconsistent interpretations. The occurrence of these extraction mistakes reduces the reported accuracy, resulting in a lower measured level of faithfulness than it actually is.

\textbf{Category 3. Counterintuitive Feature Influence}

The \textit{Narrator} is reluctant to follow the given SHAP explanation when its information contradicts widely accepted real-world common sense. For example, the SHAP table for index 11 in the \textit{Student} dataset (from baseline experiment) indicates that the absence of family support has a positive effect on the prediction of student success, which is counterintuitive to common beliefs. The \textit{Narrator} generates a narrative that describes the lack of family support as a negative factor on students' performance, as this more closely aligns with common knowledge.

\textbf{Category 4. Misleading Coherence Feedback}

This situation occurs when {the \textit{\textit{Coherence Agent}} is incorporated in the agentic system. Because its sole responsibility is to enhance the narrative’s coherence, it may provide misleading feedback that causes the \textit{Narrator} to deviate from the SHAP information. For example, the \textit{Coherence Agent} may suggest grouping similar features to improve narrative flow from one factor to another, or it may suggest reversing a feature's influence from positive to negative to better align the narrative with common sense. One real example from the \textit{Coherence Agent}'s feedback is shown in the Section~\ref{coanalysis} (case d).  

\textbf{Category 5. Generation Misalignment }

Even though the system has all the necessary and correct information, LLM can still generate narratives that are not 100\%  faithful to the given SHAP table. This explains all remaining cases not covered by the four categories above; that said, no explanation is found for this unfaithful generation from either the counterintuitive feature influence in SHAP or the misleading feedback from agents. Hence, we only allocate C5 to baseline narratives. \textit{Generation Misalignment} occurs randomly and to a large extent due to the stochasticity of LLMs.  It is one of the main reasons the agentic approach is proposed: one-time generation introduces issues related to faithfulness, even for the best LLM.

It is important to note that the categorization process is not always clean-cut. Across the three refinement rounds, the reason a narrative contains faithfulness issues may change. In other cases, even within a single round, the cause of the error can be mixed. To handle this complexity, we adopt a principle of `backward". We first examine the round 2 narrative to determine whether a clear cause is identified. If no cause is evident, we trace backwards to the round 1 feedback to determine whether the feedback provided to the \textit{Narrator} introduced any cause. When multiple categories apply, we assign the narrative to the dominant category.

\begin{figure}[!b]
    \centering
    \includegraphics[width=0.7\textwidth]{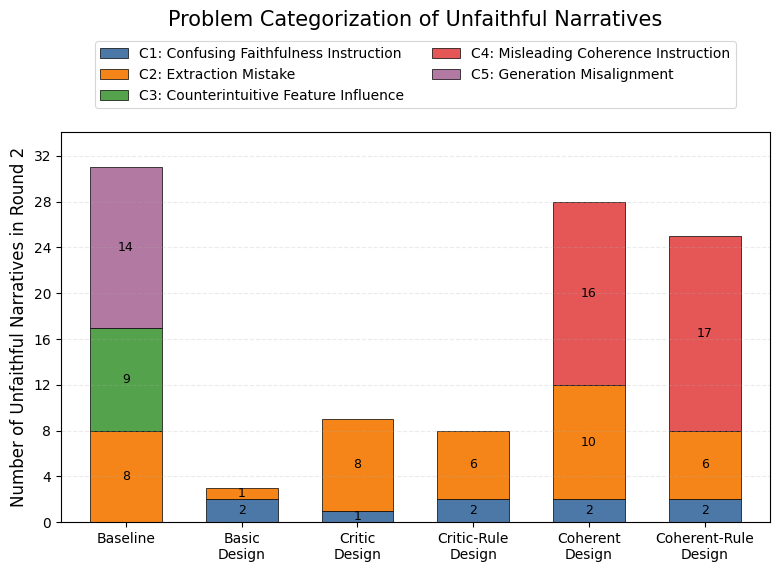}
    \caption{This bar chart shows the distribution of problem categories across five system designs (Claude Sonnet 4.5).}
    \label{errorchart}
\end{figure}

Looking into Figure~\ref{errorchart} on the distribution of the problem categories for each agentic system design, we see that the biggest problem for each agentic system is different, with \textit{C5 (Generation Misalignment)} for the baseline experiment and \textit{C4 (Misleading Coherence Instruction)} for the agentic system with the \textit{\textit{Coherence Agent}}. \textit{C2 (Extraction Mistake)} accounts for a noticeable part across all designs, which motivates us with an ensemble strategy to get a more reliable extraction (Section~\ref{ensemble}). Moreover, the two major problems (\textit{C3: Counterintuitive Feature Influence} and \textit{C5: Generation Misalignment}) in the baseline experiment are addressed in the agentic designs, demonstrating the effectiveness of the agentic approach in rectifying the flaws of one-time generation. Based solely on the reported accuracy, the \textit{Basic Design} is the best-performing across all five designs. However, since the narratives that are categorized as having \textit{Extraction Mistakes} are faithful in fact,  this detailed examination reveals that the \textit{Critic-Rule Design} and the \textit{Critic Design} are comparable with the \textit{Basic Design} in reality. 

We subsequently check all the narratives reported with extraction mistakes in the \textit{Critic Design} that are not in the \textit{Basic Design} to understand in depth why extraction mistakes happen less. For half of them, the \textit{\textit{Faithful Evaluator}} just doesn't make an extraction mistake, luckily. For the other half, we observe that the \textit{Narrator} in the \textit{Basic Design} modifies a few descriptive terms to make the importance ranking of the features clearer and easier to extract. A comparison between them (Figure~\ref{EM} and Figure~\ref{fixedEM}):

\begin{figure}[h!]
\centering
    \begin{tcolorbox}[colback=white!5, colframe=darkbluegrey, title=\textbf{Fifa: index 0 (Critic Design)}]
    \{A starting sentence with prediction score\}
    
    The most influential factor was the team failing to score any goals, which significantly reduced their chances—since goals are typically decisive in earning individual accolades. Despite having higher-than-average ball possession (64\% vs. the average 50\%), this only \colorbox{orange!20}{\textbf{slightly}} improved their odds, as possession alone doesn't guarantee impactful play. The team's 13 attempts on goal, slightly above average, provided a \colorbox{orange!20}{\textbf{minor}} boost, suggesting they created opportunities but lacked finishing. Finally, committing 22 fouls—well above the average—hurt their chances, \colorbox{orange!20}{\textbf{likely}} reflecting a lack of discipline that may have disrupted their rhythm.
    
    \{A conclusion sentence\}
    \end{tcolorbox} 

\caption{An Example of \textit{Extraction Mistake} on \textit{Rank}. Due to the subtle semantic differences of descriptive terms, the task of extracting the correct feature importance is difficult for the \textit{Faithful Evaluator}. }
\label{EM}
\end{figure}

\begin{figure}[h!]
\centering
    \begin{tcolorbox}[colback=white!5, colframe=darkbluegrey, title=\textbf{Fifa: index 0 (Basic Design)}]
    \{A starting sentence with prediction score\}
    
    The most influential factor was the team failing to score any goals, which significantly reduced their chances—since goals are typically decisive in earning individual accolades. Ball possession at 64\%, well above the average of 50\%, was the \colorbox{green!20}{\textbf{second most impactful}} factor and slightly improved their odds, though possession alone doesn't guarantee the creative or clinical play needed for such recognition. The team's 13 attempts on goal, \colorbox{green!20}{\textbf{marginally}} above average, provided a modest boost as the third most important feature, suggesting they created opportunities but lacked the finishing quality to convert them. Committing 22 fouls—considerably above the average—was \colorbox{green!20}{\textbf{the fourth factor}} and negatively affected their chances, likely reflecting poor discipline that may have disrupted their flow and created an unfavorable impression.
    
    \{A conclusion sentence\}
    \end{tcolorbox}
\caption{No \textit{Extraction Mistake} on \textit{Rank}. The \textit{Basic Design} improves the clarity of the descriptive terms, making the extraction task easier. }
\label{fixedEM}
\end{figure}

The narrative from the \textit{Critic Design} is arguably faithful because the rank order of the four features totally corresponds to the SHAP table, while the adverbs don't explicitly violate. Nevertheless, in the \textit{Basic Design}, the \textit{Narrator} modifies the descriptive terms to clarify feature importance, thereby substantially reducing extraction mistakes. To summarize, the \textit{\textit{Faithful Critic}} provides overly explicit feedback focused on sentence reordering and additionally includes a constraint prohibiting changes to any other parts of the narrative. This restriction limits the \textit{Narrator}’s ability to consider alternative solutions. In contrast, more general feedback by the \textit{\textit{Faithful Evaluator}} imposes no such constraints. Hence, \textit{Narrator} can flexibly revise those inappropriate elements, thereby facilitating correct extraction.

In conclusion, the Claude Sonnet 4.5-based \textit{Basic Design} outperforms the \textit{Critic Design} depending on several underlying factors. Fundamentally, the \textit{Basic Design} is not necessarily superior to the \textit{Critic Design} or the \textit{Critic-Rule Design}, if we consider those extraction mistakes. Beyond this, linguistic nuances inject difficulty and uncertainty into the extraction task. Third, the Claude-based \textit{Basic Design} demonstrates the ability to revise narratives effectively even with vague feedback, a capability not consistently observed across all LLMs. 

\subsection{Faithfulness Results across LLMs}\label{originalall}

\begin{table*}[b!]
\centering
\caption{Faithfulness results for different agentic system designs based on different LLMs. Arrows indicate the progression of the rank, sign, and value accuracy from Round 0 to Round 2. The last column shows the overall accuracy in round 2 for each configuration.}
\resizebox{\textwidth}{!}{\small
    \setlength{\tabcolsep}{4pt}
    \begin{tabular}{l l ccc ccc ccc c}
    \toprule
    \multirow{2}{*}{\textbf{Designs}} & \multirow{2}{*}{\textbf{LLMs}} &
    \multicolumn{3}{c}{\textbf{Faithfulness}} & \multirow{2}{*}{\textbf{Overall (round 2)}} \\
    \cmidrule(lr){3-5}
    & &
    \textbf{{RA}} & \textbf{{SA}} & \textbf{{VA}} & \\
    \midrule
    
    \multirow{5}{*}{\textbf{\makecell[l]{Basic Design}}}
    & \underline{Claude Sonnet 4.5}
    & 0.905$\rightarrow$0.970$\rightarrow$\underline{0.990}
    & 0.957$\rightarrow$0.997$\rightarrow$\underline{1.000}
    & 0.990$\rightarrow$0.995$\rightarrow$0.997
    & \underline{0.996} \\
    & Mistral Medium 3.1
    & 0.898$\rightarrow$0.870$\rightarrow$0.887
    & 0.950$\rightarrow$0.965$\rightarrow$0.970
    & 0.990$\rightarrow$0.995$\rightarrow$0.995
    & 0.951 \\
    & GPT-5
    & 0.898$\rightarrow$0.945$\rightarrow$0.972
    & 0.965$\rightarrow$0.995$\rightarrow$0.997
    & 0.993$\rightarrow$0.995$\rightarrow$\underline{1.000}
    & 0.990 \\
    & Llama 3.3 70B Instruct
    & 0.857$\rightarrow$0.875$\rightarrow$0.863
    & 0.940$\rightarrow$0.960$\rightarrow$0.975
    & 0.993$\rightarrow$0.997$\rightarrow$\underline{1.000}
    & 0.946 \\
    & DeepSeek-V3.2-Exp
    & 0.905$\rightarrow$0.943$\rightarrow$0.940
    & 0.960$\rightarrow$0.988$\rightarrow$0.980
    & 0.992$\rightarrow$0.988$\rightarrow$0.993
    & 0.971 \\
    \midrule
    
    \multirow{5}{*}{\textbf{\makecell[l]{Critic Design}}}
    & Claude Sonnet 4.5
    & 0.905$\rightarrow$0.950$\rightarrow$0.960
    & 0.960$\rightarrow$0.988$\rightarrow$0.993
    & 0.990$\rightarrow$0.995$\rightarrow$0.997
    & 0.983 \\
    & Mistral Medium 3.1
    & 0.897$\rightarrow$0.963$\rightarrow$0.975
    & 0.952$\rightarrow$0.990$\rightarrow$0.993
    & 0.990$\rightarrow$0.997$\rightarrow$0.997
    & 0.988 \\
    & GPT-5
    & 0.902$\rightarrow$0.940$\rightarrow$0.948
    & 0.965$\rightarrow$0.997$\rightarrow$\underline{1.000}
    & 0.993$\rightarrow$0.992$\rightarrow$\underline{1.000}
    & 0.982 \\
    & Llama 3.3 70B Instruct
    & 0.875$\rightarrow$0.900$\rightarrow$0.927
    & 0.950$\rightarrow$0.950$\rightarrow$0.980
    & 0.993$\rightarrow$0.995$\rightarrow$0.997
    & 0.968 \\
    & \underline{DeepSeek-V3.2-Exp}
    & 0.910$\rightarrow$0.988$\rightarrow$\underline{0.988}
    & 0.950$\rightarrow$0.988$\rightarrow$0.985
    & 0.995$\rightarrow$0.995$\rightarrow$0.997
    & \underline{0.990} \\
    \midrule

    \multirow{5}{*}{\textbf{\makecell[l]{Critic-Rule Design}}}
    & Claude Sonnet 4.5
    & 0.905$\rightarrow$0.960$\rightarrow$0.960
    & 0.957$\rightarrow$0.993$\rightarrow$0.997
    & 0.990$\rightarrow$0.995$\rightarrow$0.997
    & 0.985 \\
    & \underline{Mistral Medium 3.1}
    & 0.907$\rightarrow$0.955$\rightarrow$\underline{0.980}
    & 0.952$\rightarrow$0.985$\rightarrow$0.990
    & 0.990$\rightarrow$0.995$\rightarrow$0.997
    & \underline{0.989} \\
    & \underline{GPT-5}
    & 0.885$\rightarrow$0.957$\rightarrow$0.968
    & 0.963$\rightarrow$0.997$\rightarrow$\underline{1.000}
    & 0.993$\rightarrow$0.997$\rightarrow$\underline{1.000}
    & \underline{0.989} \\
    &Llama 3.3 70B Instruct
    & 0.847$\rightarrow$0.912$\rightarrow$0.930
    & 0.940$\rightarrow$0.970$\rightarrow$0.990
    & 0.993$\rightarrow$0.993$\rightarrow$\underline{1.000}
    & 0.973 \\
    & DeepSeek-V3.2-Exp
    & 0.912$\rightarrow$0.970$\rightarrow$0.978
    & 0.960$\rightarrow$0.980$\rightarrow$0.980
    & 0.992$\rightarrow$0.997$\rightarrow$\underline{1.000}
    & 0.986 \\
    \midrule

    \multirow{5}{*}{\textbf{\makecell[l]{Coherent Design}}}
    & Claude Sonnet 4.5
    & 0.905$\rightarrow$0.915$\rightarrow$0.860
    & 0.957$\rightarrow$0.982$\rightarrow$0.980
    & 0.990$\rightarrow$0.993$\rightarrow$0.997
    & 0.946 \\
    & Mistral Medium 3.1
    & 0.897$\rightarrow$0.790$\rightarrow$0.748
    & 0.952$\rightarrow$0.975$\rightarrow$0.977
    & 0.993$\rightarrow$0.995$\rightarrow$0.995
    & 0.907 \\
    & GPT-5
    & 0.893$\rightarrow$0.897$\rightarrow$0.875
    & 0.963$\rightarrow$0.993$\rightarrow$\underline{0.990}
    & 0.993$\rightarrow$0.992$\rightarrow$0.997
    & 0.954 \\
    & Llama 3.3 70B Instruct
    & 0.872$\rightarrow$0.768$\rightarrow$0.820
    & 0.953$\rightarrow$0.972$\rightarrow$0.960
    & 0.993$\rightarrow$0.985$\rightarrow$\underline{1.000}
    & 0.927 \\
    & \underline{DeepSeek-V3.2-Exp}
    & 0.922$\rightarrow$0.945$\rightarrow$\underline{0.922}
    & 0.950$\rightarrow$0.967$\rightarrow$0.982
    & 0.992$\rightarrow$0.993$\rightarrow$0.997
    & \underline{0.967} \\
    \midrule
    
    \multirow{5}{*}{\textbf{\makecell[l]{Coherent-Rule Design}}}
    & \underline{Claude Sonnet 4.5}
    & 0.905$\rightarrow$0.900$\rightarrow$\underline{0.878}
    & 0.957$\rightarrow$0.985$\rightarrow$0.985
    & 0.990$\rightarrow$0.990$\rightarrow$0.997
    & \underline{0.953} \\
    & Mistral Medium 3.1
    & 0.903$\rightarrow$0.765$\rightarrow$0.770
    & 0.952$\rightarrow$0.975$\rightarrow$0.980
    & 0.988$\rightarrow$0.995$\rightarrow$0.997
    & 0.916 \\
    & \underline{GPT-5}
    & 0.885$\rightarrow$0.922$\rightarrow$0.870
    & 0.963$\rightarrow$0.993$\rightarrow$\underline{0.988}
    & 0.993$\rightarrow$0.997$\rightarrow$\underline{1.000}
    & \underline{0.953} \\
    & Llama 3.3 70B Instruct
    & 0.865$\rightarrow$0.800$\rightarrow$0.730
    & 0.952$\rightarrow$0.972$\rightarrow$0.940
    & 0.993$\rightarrow$0.982$\rightarrow$\underline{1.000}
    & 0.890 \\
    & DeepSeek-V3.2-Exp
    & 0.907$\rightarrow$0.938$\rightarrow$0.882
    & 0.950$\rightarrow$0.965$\rightarrow$0.948
    & 0.992$\rightarrow$0.997$\rightarrow$0.995
    & 0.942 \\
    
    \bottomrule
    \end{tabular}
}
\footnotesize{Note. For each agentic design, the highest rank accuracy, sign accuracy, and value accuracy are underlined. The LLM with the highest overall round-2 accuracy is also underlined. It is notable that different LLMs extract different accuracy on the same baseline narratives; for each LLM, extraction can also be unstable across runs (see appendix~\ref{llmexapp}). }
\label{llmstable}
\end{table*}

Table~\ref{llmstable} indicates the faithfulness results of all agentic designs across five LLMs. Among all 25 combinations of LLMs and agentic designs, the Claude Sonnet 4.5-based \textit{Basic Design} ranks highest, achieving an overall round-2 accuracy of 0.996. When aggregating performance across five agentic system designs, GPT-5, Claude Sonnet 4.5, and DeepSeek-V3.2-Exp exhibit comparable performance. Mistral Medium 3.1 ranks next, whereas Llama 3.3 70B Instruct is noticeably the worst. For each specific design, the LLM that performs the best varies somewhat. For example, the winner for both the \textit{Basic Design} and the \textit{Coherent-Rule Design} is Claude Sonnet 4.5. DeepSeek-V3.2-Exp wins for both the \textit{Critic Design} and the \textit{Coherent Design}, and the best LLMs on the \textit{Critic-Rule Design} are Mistral Medium 3.1 and GPT-5. Llama 3.3 70B Instruct never ranks in the top for any design. 

Since the \textit{Critic-Rule Design} is the overall best design across all five LLMs, we plot its results in a bar chart to have a direct comparison (Figure \ref{llmsbar}). Mistral Medium 3.1 and DeepSeek-V3.2-Exp improve the rank accuracy comparably well, while GPT-5 and Claude Sonnet 4.5 improve the sign accuracy the most. All LLMs steadily improve value accuracy, partly due to the already high baseline accuracy. We also observe that, although the same batch of baseline narratives is used to provide a consistent starting point, the reported baseline accuracies across LLMs differ, particularly in rank accuracy. This further reflects the extraction mistakes introduced differently across LLMs, motivating our ensemble strategy.

\begin{figure}[h!]
    \centering
    \includegraphics[width=1.0\textwidth]{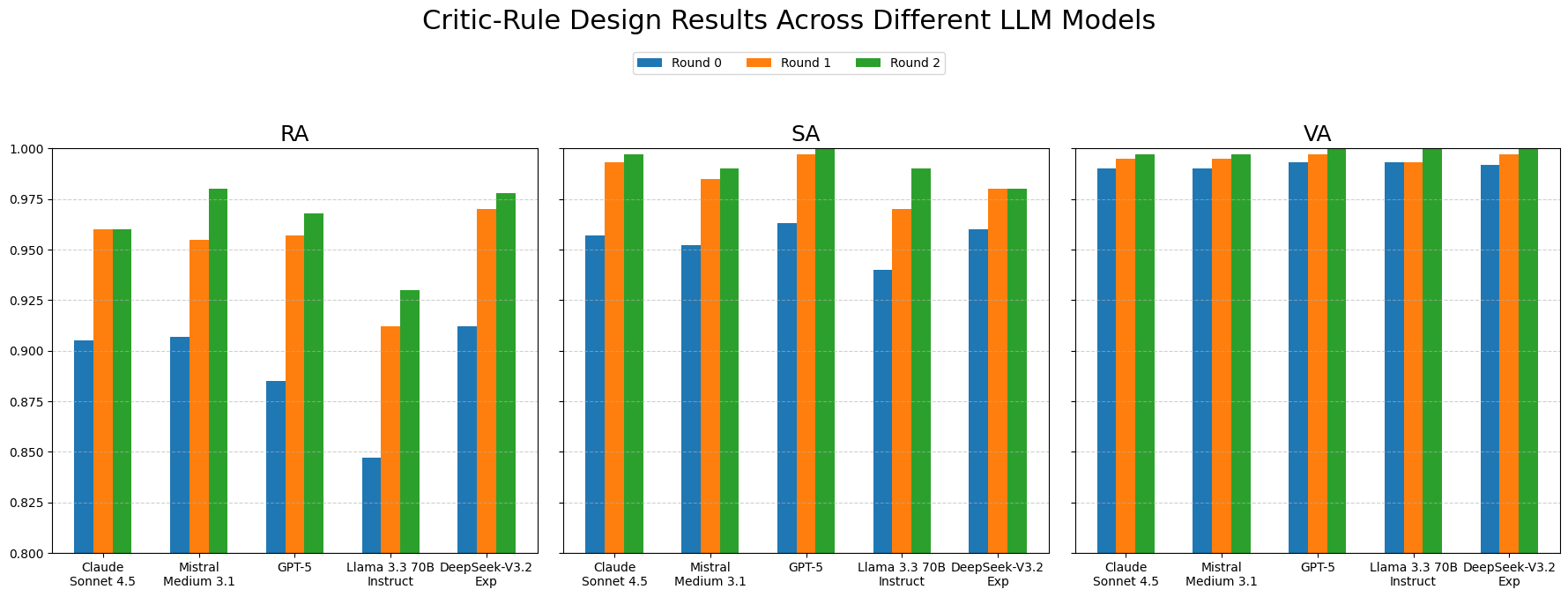}
    \caption{This bar chart presents the three accuracies (\textit{RA}: rank accuracy,\textit{ SA}: sign accuracy, and \textit{VA}: value accuracy) among five LLMs on the \textit{Critic-Rule Design}.}
    \label{llmsbar}
\end{figure}

\begin{table}[b!]
\centering
\footnotesize
\caption{The problem categorization on the round-2 unfaithful narratives reported by the \textit{Basic Design} and the \textit{Critic Design} based on DeepSeek-V3.2-Exp. }
\begin{tabular}{l|c|l|c|l}
\hline
\textbf{DeepSeek-V3.2-Exp} & \multicolumn{2}{c}{\textbf{\textit{Basic Design}}} & \multicolumn{2}{c}{\textbf{\textit{Critic Design}}} \\ \hline
\textbf{Problem Categorization} & \textbf{number} & \textbf{instance} & \textbf{number} & \textbf{instance} \\ \hline
C1: Confusing Faithfulness Instruction & 10 & \begin{tabular}[c]{@{}l@{}}diabetes: index 2,4,7,17\\ stroke: index 13\\ student: index 1, 12, 14\\ credit: index 6,10\end{tabular} & 1 & student: index 14 \\ \hline
C2: Extraction Mistake & 3 & \begin{tabular}[c]{@{}l@{}}stroke: index 2\\ student: index 5, 18\end{tabular} & 3 & \begin{tabular}[c]{@{}l@{}}stroke: index 2\\ student: index 3, 5\end{tabular} \\ \hline
C3: Counterintuitive Feature Influence & 1 & student: index 16 & 1 & student: index 16 \\ \hline
C4: Misleading Coherence Instruction &  &  &  &  \\ \hline
C5: Generation Misalignment &  &  &  &  \\ \hline
\textbf{In Total} & \textbf{14} &  & \textbf{5} &  \\ \hline
\end{tabular}
\label{basicvscritic}
\end{table}

We further investigate whether different LLMs exhibit distinct optimal agentic designs and analyze the underlying reasons for these differences. The optimal design for Claude Sonnet 4.5 is the \textit{Basic Design}, whereas the other LLMs perform the best either on the \textit{Critic Design} or the \textit{Critic-Rule Design}. GPT-5 performs almost as well for the \textit{Basic Design} and the \textit{Critic-Rule Design}. 

We use DeepSeek-V3.2-Exp as a representative to examine why it works better in the \textit{Critic Design} than the \textit{Basic Design}. Overall, the DeepSeek-V3.2-Exp-based \textit{Basic Design} system produces 14 unfaithful narratives by round 2, whereas the \textit{Critic Design} outputs 5. Among the 14 unfaithful cases under \textit{Basic Design}, 10 are attributed to unclear feedback from the \textit{Faithful Evaluator (C1)} . These cases are resolved in the \textit{Critic Design} because of the explicit instructions provided by the \textit{Faithful Critic}. Of the remaining cases in \textit{Basic Design}, 3 result from extraction mistakes (\textit{C2}), and 1 stems from counterintuitive feature influence (\textit{C3}). In contrast, for the 5 unfaithful narratives produced by the \textit{Critic Design}, the distribution of problem types is as follows: \textit{C1} (1 narrative), \textit{C2} (3 narratives), and \textit{C3} (1 narrative) (Table~\ref{basicvscritic}). In conclusion, the design incorporated with\textit{ the \textit{Faithful Critic}} or \textit{the \textit{Faithful Critic (Rule)} } tends to outperform a design with only the \textit{Faithful Evaluator}, as the \textit{Narrator} can't fix errors as efficiently without a clear, directional revision guidance (Deepseek-V3.2-Exp, Mistral Medium 3.1 and Llama 3.3 70B Instruct). In addition, differences in the optimal agentic designs also imply the superior performance of Claude Sonnet 4.5 and GPT-5: the \textit{Narrator} is able to infer the appropriate revision based on the prompt, which contains the original SHAP information. That being said, even with the vague \textit{\textit{Faithful Evaluator}} feedback, Claude Sonnet 4.5 and GPT-5, possibly due to their capability of taking a larger context window, may trace back to the lengthy prompt and get the relevant information for their revision.

\subsection{Faithfulness Results for Ensemble Method}\label{ensemble}
Table~\ref{llmstableensemble} presents the ensemble experiments' results. Across all 25 experiments using the ensemble method, most show performance improvements, with exception cases marked in red.  Llama 3.3 70B Instruct, which performs worst in the original experiments, shows the largest improvement and reaches a level comparable to DeepSeek-V3.2-Exp. Most notably, the \textit{Basic Design} design reaches an accuracy of 0.999, with only a single narrative remaining unfaithful (\textit{Credit}: index 10, C1.1). Considering the overall round-2 accuracy, only five experiments show decreases: Mistral Medium 3.1 and GPT-5 under the \textit{Critic-Rule Design}, and DeepSeek-V3.2-Exp under the \textit{Critic Design}, the \textit{Critic-Rule Design} and the \textit{Coherent Design}. As a complementary indicator, the number of unfaithful narratives in round 2 largely mirrors these accuracy trends. For Claude Sonnet 4.5, GPT-5, and Llama 3.3 70B Instruct, the ensemble experiments consistently reduce the number of unfaithful narratives across all agentic designs, reflecting the effectiveness of the ensemble strategy. By contrast, DeepSeek-V3.2-Exp-based agentic systems using an ensemble approach exhibit a clear failure. This suggests that DeepSeek-V3.2-Exp performs better on the extraction task, with its strengths diluted by the inclusion of lower-performing LLMs in the ensemble method. This hypothesis is further validated by the manual analysis presented in the following section.

\begin{table*}[h!]
\centering
\caption{This table compares the round-2 accuracy obtained from the original experiments (o) and the ensemble experiments (e). Numbers are highlighted in red if the ensemble experiment's result is worse than that of the original experiment. The final column reports the number of remaining unfaithful narratives from each. }
\footnotesize
    \setlength{\tabcolsep}{4pt}
    \begin{tabular}{l l ccc ccc ccc c c}
    \toprule
    \multirow{2}{*}{\textbf{LLMs}} & \multirow{2}{*}{\textbf{Designs}} &
    \multicolumn{3}{c}{\textbf{Faithfulness}} & \makecell[c]{\textbf{Overall} \\ \textbf{(round 2)}} & \makecell[c]{\textbf{No. Unf. Narr.} \\ \textbf{(round 2)}}\\ %
    \cmidrule(lr){3-5}
    & &
    {\textbf{RA(o)$\vert$RA(e)}} & {\textbf{SA(o)$\vert$SA(e)}} & {\textbf{VA(o)$\vert$VA(e)}}& \textbf{(o)$\vert$(e)}  & \textbf{(o)$\vert$(e)} \\
    \midrule
    
    \multirow{5}{*}{\textbf{\makecell[l]{Claude Sonnet 4.5}}}
    & Basic Design
    & 0.990$\vert$1.000
    & 1.000$\vert$1.000
    & 0.997$\vert$0.997
    & 0.996$\vert$\underline{\underline{0.999}}
    & 3$\vert$1 \\ 
    & Critic Design
    & 0.960$\vert$0.970
    & 0.993$\vert$0.997
    & 0.997$\vert$0.997
    & 0.983$\vert$0.988 
    &9$\vert$6 \\
    & Critic-Rule Design
    & 0.960$\vert$0.970
    & 0.997$\vert$0.997
    & 0.997$\vert$0.997
    & 0.985$\vert$0.988 
    & 8$\vert$6 \\
    & Coherent Design
    & 0.860$\vert$0.872
    & 0.980$\vert$0.990
    & 0.997$\vert$1.000
    & 0.946$\vert$0.954 
    & 28$\vert$23 \\
    & Coherent-Rule Design
    & 0.878$\vert$0.892
    & 0.985$\vert$0.993
    & 0.997$\vert$1.000
    & 0.953$\vert$0.962 
    & 25$\vert$20 \\
    \midrule
    
    \multirow{5}{*}{\textbf{\makecell[l]{Mistral Medium 3.1}}}
    & Basic Design
    & 0.887$\vert$0.955
    & 0.970$\vert$1.000
    & 0.995$\vert$1.000
    & 0.951$\vert$0.985 
    &21$\vert$7 \\
    & Critic Design
    & 0.975$\vert$0.990
    & 0.993$\vert$0.997
    & 0.997$\vert$1.000
    & 0.988$\vert$0.996 
    & 8$\vert$3 \\
    & Critic-Rule Design
    & 0.980$\vert$\textcolor{red}{0.968}
    & 0.990$\vert$0.990
    & 0.997$\vert$\textcolor{red}{0.995}
    & 0.989$\vert$\textcolor{red}{0.984}  
    & 5$\vert$\textcolor{red}{8} \\
    & Coherent Design
    & 0.748$\vert$0.765
    & 0.977$\vert$0.990
    & 0.995$\vert$1.000
    & 0.907$\vert$0.918 
    & 47$\vert$40 \\
    & Coherent-Rule Design
    & 0.770$\vert$0.785
    & 0.980$\vert$0.995
    & 0.997$\vert$\textcolor{red}{0.995}
    & 0.916$\vert$0.925 
    & 41$\vert$39 \\
    \midrule
    
    \multirow{5}{*}{\textbf{\makecell[l]{GPT-5}}}
    & Basic Design
    & 0.972$\vert$0.982
    & 0.997$\vert$0.997
    & 1.000$\vert$1.000
    & 0.990$\vert$0.993 
    & 6$\vert$4 \\
    & Critic Design
    & 0.948$\vert$0.968
    & 1.000$\vert$\textcolor{red}{0.997}
    & 1.000$\vert$1.000
    & 0.982$\vert$0.988 
    & 9$\vert$7 \\
    & Critic-Rule Design
    & 0.968$\vert$0.970
    & 1.000$\vert$\textcolor{red}{0.997}
    & 1.000$\vert$\textcolor{red}{0.997}
    & 0.989$\vert$\textcolor{red}{0.988} 
    & 10$\vert$8 \\
    & Coherent Design
    & 0.875$\vert$0.908
    & 0.990$\vert$0.997
    & 0.997$\vert$1.000
    & 0.954$\vert$0.968 
    & 25$\vert$17 \\
    & Coherent-Rule Design
    & 0.870$\vert$0.950
    & 0.988$\vert$0.988
    & 1.000$\vert$\textcolor{red}{0.995}
    & 0.953$\vert$0.978 
    & 24$\vert$14 \\
    \midrule
    
    \multirow{5}{*}{\textbf{\makecell[l]{Llama 3.3 70B Instruct}}}
    & Basic Design
    & 0.863$\vert$0.927
    & 0.975$\vert$0.985
    & 1.000$\vert$\textcolor{red}{0.997}
    & 0.946$\vert$0.970 

    & 23$\vert$15 \\ 
    & Critic Design
    & 0.927$\vert$0.963
    & 0.980$\vert$0.997
    & 0.997$\vert$0.997
    & 0.968$\vert$0.986 
    & 13$\vert$7 \\
    & Critic-Rule Design
    & 0.930$\vert$0.965
    & 0.990$\vert$0.997
    & 1.000$\vert$\textcolor{red}{0.997}
    & 0.973$\vert$0.987 
    & 12$\vert$8 \\
    & Coherent Design
    & 0.820$\vert$0.872
    & 0.960$\vert$0.985
    & 1.000$\vert$1.000
    & 0.927$\vert$0.953 
    & 28$\vert$26 \\
    & Coherent-Rule Design
    & 0.730$\vert$0.863
    & 0.940$\vert$0.975
    & 1.000$\vert$1.000
    & 0.890$\vert$0.946 
    & 35$\vert$27 \\
    \midrule
    
    \multirow{5}{*}{\textbf{\makecell[l]{DeepSeek-V3.2-Exp}}}
    & Basic Design
    & 0.940$\vert$\textcolor{red}{0.938}
    & 0.980$\vert$0.993
    & 0.993$\vert$\textcolor{red}{0.992}
    & 0.971$\vert$0.974 

    & 14$\vert$\textcolor{red}{15} \\ %
    & Critic Design
    & 0.988$\vert$\textcolor{red}{0.963}
    & 0.985$\vert$0.993
    & 0.997$\vert$1.000
    & 0.990$\vert$\textcolor{red}{0.985} 
    & 5$\vert$\textcolor{red}{9} \\
    & Critic-Rule Design
    & 0.978$\vert$\textcolor{red}{0.957}
    & 0.980$\vert$0.997
    & 1.000$\vert$\textcolor{red}{0.997}
    & 0.986$\vert$\textcolor{red}{0.984} 
    & 7$\vert$7 \\
    & Coherent Design
    & 0.922$\vert$\textcolor{red}{0.885}
    & 0.982$\vert$0.985
    & 0.997$\vert$0.997
    & 0.967$\vert$\textcolor{red}{0.956} 
    & 21$\vert$\textcolor{red}{23} \\
    & Coherent-Rule Design
    & 0.882$\vert$\textcolor{red}{0.840}
    & 0.948$\vert$0.993
    & 0.995$\vert$0.995
    & 0.942$\vert$0.942 
    & 24$\vert$\textcolor{red}{32} \\
    
    \bottomrule
    \end{tabular}
\label{llmstableensemble}
\end{table*}

\subsubsection{Problem Categorization of Unfaithful Narratives (Ensemble Method)}\label{categorization_en}

We categorize the round-2 unfaithful narratives from the ensemble experiments to examine whether the error patterns differ from those observed in the original experiments. For consistency, we again focus on results obtained with Claude Sonnet 4.5, as shown in Figure~\ref{errorchart_compare}. A first observation is that the number of \textit{C2 (Extraction Mistake)} indeed decreases; however, it doesn't disappear. Issues from the \textit{Coherence Agent}'s misleading suggestion (\textit{C4}) and confusing faithfulness-related instructions (\textit{C1}) still occur, as expected. 

The primary reason why extraction mistakes remain is that majority voting does not guarantee a correct final extraction, which just mitigates the incorrect one-time extraction from a single LLM. For narratives with semantic similarity in descriptive terms, accurate extraction remains challenging even for strong LLMs. Consequently, when only one or two models extract the information correctly, the majority vote may still result in an incorrect extraction, let alone a case when it is too challenging, and no LLM succeeds. Among the 18 failed cases from all five designs, 10 of them result in a 2:3 split, 4 in a 1:4 split, and 4 cases in a 0:5 split. There is not always a single LLM that consistently produces the correct extraction, with the LLama 3.3 70B Instruct model erring most often. Nevertheless, DeepSeek-V3.2-Exp appears to be the most capable at handling these difficult cases, producing the correct extraction in 13 of these instances. This observation helps explain why DeepSeek-V3.2-Exp exhibits lower accuracy in the ensemble experiments: the increased number of extraction errors introduced by other voters harms the ensemble results.

In conclusion, the ensemble method indeed improve performance for most cases. However, the extraction mistakes cannot be entirely eliminated due to the inherent difficulty of the extraction task in linguistically complex texts. Moreover, when an LLM, such as DeepSeek-V3.2-Exp, is already sufficiently strong in the extraction task, the ensemble method degrades the agentic system performance.

\begin{figure}[hb!]
    \centering
    \includegraphics[width=0.7\textwidth]{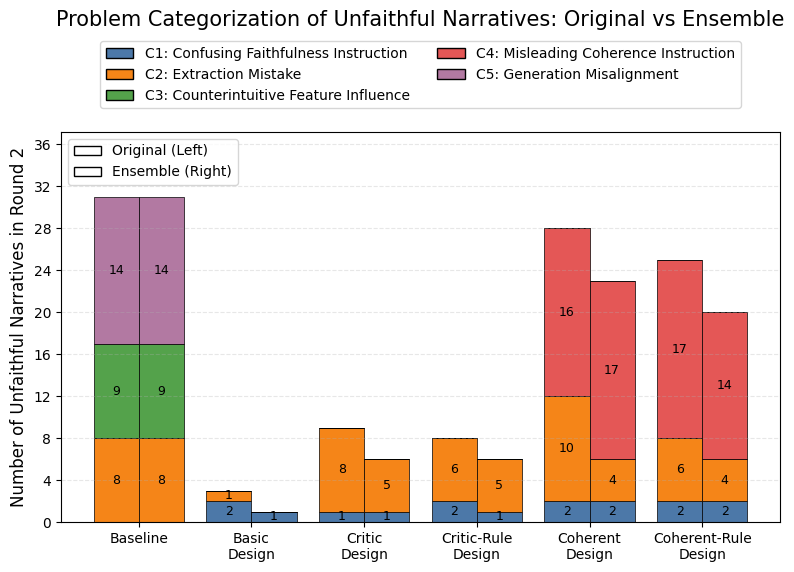}
    \caption{A comparison between the problem categorization: original and ensemble experiments (Claude Sonnet 4.5). }
    \label{errorchart_compare}
\end{figure}

\subsection{Additional Results on Key Concepts}\label{addition}

We decide to choose the best agentic system design as the representative to evaluate key concepts, such as the influence of the number of iterations, the number of features included in the narrative, etc. However, as shown in Table~\ref{llmstableensemble}, no single design consistently outperformed others across all LLMs, and ensemble experiments do not always perform better than the original experiments. We hence propose to determine the representative design based on the frequency of ``wins" in overall accuracy (round 2). In total, the \textit{Critic Design }and the \textit{Critic-Rule Design} reach the top rank five times each, outperforming the \textit{Basic Design} (four wins). Between these two, the \textit{Critic-Rule Design} is superior to the \textit{Critic Design} with a 5:3 ratio (with two ties). Furthermore, for three out of five LLMs, the original \textit{Critic-Rule} outperforms the ensemble version (Mistral Medium 3.1, GPT-5, and DeepSeek-V3.2-Exp) (see Table~\ref{benchmarkchoice} in Appendix~\ref{benchmarkselection}). Given that the original experiments are also significantly more efficient on resource usage, we finally select the original \textit{Critic-Rule Design} as our benchmark for further experimentation. 

The following key concepts are of interest:

\begin{itemize}
    \item \textbf{Iteration Round}. How does the number of iteration rounds impact the result?
    \item \textbf{The Number of Features}. How does the number of features included in the narrative impact the result?
    \item \textbf{No-Baseline Setting}. What if we don't use baseline narratives? How does this impact the result?
    \item \textbf{Different Baseline Narratives}. How do the best and the worst baseline narratives impact the result?
\end{itemize}

\textbf{Iteration Round} We change the maximum round of improvement to 10 in this experiment to see how this impacts the final accuracy (Table~\ref{round=10}, Figure~\ref{round=10plot}). There is a rapid improvement of rank, sign, and value accuracy in the first two rounds across all LLMs. Thereafter, performance reaches a plateau from round 2 to round 9 for most LLMs, with minor fluctuations. Deepseek-V3.2-Exp performs the best across all five LLMs, with an overall round-9 accuracy of 0.998. Llama 3.3 70B shows a slightly different pattern. It exhibits a slight decline in rank and sign accuracy during the middle stages of the iteration, followed by a gradual improvement until reaching a level comparable to GPT-5 and Claude Sonnet 4.5. The best performance is achieved after six rounds of iteration, more slowly than other LLMs. We also examine the underlying reasons for the round-9 unfaithful narratives and categorize them into the five problem categories. The primary reason for every LLM is extraction mistakes caused by linguistic ambiguity (Table~\ref{round=10table}). For GPT-5 and Claude Sonnet 4.5, two instances remain unfixed due to the confusing instructions from the \textit{Faithful Critic}, as elaborated in section~\ref{categorization} and Figure~\ref{C1.2}. 

To summarize, most LLMs resolve faithfulness-related issues rapidly, typically within two rounds of iteration, whereas Llama 3.3 70B is the least efficient, requiring six rounds. The main reason none of the experiments achieve 100\% accuracy by round 9 is that the extraction mistakes each LLM causes on its own are unsolvable due to its capability limitations.

\textbf{The Number of Features} We increase the number of features included in each narrative from four to eight to evaluate the robustness of the agentic system in improving faithfulness. As in the previous setting, we use DeepSeek-V3-0324 to generate 100 baseline narratives containing eight features and adopt them as the new baseline for this experiment. The results show that round-2 accuracy improves across all five LLMs (Table~\ref{4additional}). Both round-0 and round-2 accuracies exhibit a noticeable decline compared to the original experiment with four features. This outcome is expected, as increasing the number of features inherently raises the complexity of both generation and extraction. Notably, DeepSeek-V3.2-Exp performs particularly well under this setting, with its round-2 accuracy decreasing by only 0.001 compared to the original experiment's results.

\textbf{No-Baseline Setting} We further examine whether using baseline narratives generated by DeepSeek-V3-0324 constrains or enhances performance in the original experiments (Table~\ref{4additional}). Under this setting, DeepSeek-V3.2-Exp achieves the best results, reporting an overall round-2 accuracy of 0.999 (after manual inspection, there is an extraction mistake occurring; this is thus a 100\% faithful result). Claude Sonnet 4.5 and GPT-5 follow, with round-2 accuracies of 0.987 and 0.984, respectively. Llama 3.3 70B Instruct remains the lowest-performing model. When comparing the no-baseline results to original results, DeepSeek-V3.2-Exp is the only LLM whose overall round-2 accuracy significantly improves under the no-baseline setting. This indicates that the original baseline narratives constrain DeepSeek-V3.2-Exp's performance. On the contrary, the remaining models either experience performance declines (Mistral Medium 3.1, GPT-5, and Llama 3.3 70B Instruct) or show negligible improvement, possibly due to variance (Claude Sonnet 4.5) under a no-baseline setting. GPT-5 exhibits a distinct pattern: although it begins with its own round-0 narratives that achieve higher initial accuracy than in the original experiment, it reaches a plateau as early as round 1, resulting in slightly lower round-2 accuracy than in the original result. 

\begin{table*}[h]
\centering
\caption{The round=10 experiments' results across different LLMs (\textit{Critic-Rule Design})}
\footnotesize
    \setlength{\tabcolsep}{4pt}
    \begin{tabular}{l l c c}
    \toprule
    \textbf{LLMs} & \textbf{Faithfulness} & \textbf{Accuracies from Round 0 to Round 9} & \textbf{Overall (round 9}) \\ %
    \midrule

    \multirow{3}{*}{\textbf{Claude Sonnet 4.5}} 
    & RA & 0.905$\rightarrow$0.960$\rightarrow$0.960$\rightarrow$0.955$\rightarrow$0.960$\rightarrow$0.958$\rightarrow$0.963$\rightarrow$0.958$\rightarrow$0.963$\rightarrow$0.958 &  \multirow{3}{*}{0.984} \\  %
    & SA & 0.958$\rightarrow$0.993$\rightarrow$0.993$\rightarrow$0.993$\rightarrow$0.993$\rightarrow$0.995$\rightarrow$0.995$\rightarrow$0.995$\rightarrow$0.995$\rightarrow$0.995 & \\ 
    & VA & 0.990$\rightarrow$0.998$\rightarrow$0.998$\rightarrow$0.998$\rightarrow$0.998$\rightarrow$1.000$\rightarrow$1.000$\rightarrow$1.000$\rightarrow$1.000$\rightarrow$1.000 & \\
    \midrule    
    
    \multirow{3}{*}{\textbf{Mistral Medium 3.1}}
    & RA & 0.898$\rightarrow$0.943$\rightarrow$0.983$\rightarrow$0.985$\rightarrow$0.985$\rightarrow$0.985$\rightarrow$0.985$\rightarrow$0.990$\rightarrow$0.990$\rightarrow$0.990 & \multirow{3}{*}{0.995} \\  %
    & SA & 0.953$\rightarrow$0.985$\rightarrow$0.990$\rightarrow$0.990$\rightarrow$0.995$\rightarrow$0.998$\rightarrow$0.998$\rightarrow$0.998$\rightarrow$0.998$\rightarrow$0.998 & \\ 
    & VA & 0.993$\rightarrow$0.993$\rightarrow$0.998$\rightarrow$0.998$\rightarrow$0.998$\rightarrow$0.998$\rightarrow$0.998$\rightarrow$0.998$\rightarrow$0.998$\rightarrow$0.998 & \\
    \midrule   
    
    \multirow{3}{*}{\textbf{GPT-5}}
    & RA & 0.890$\rightarrow$0.938$\rightarrow$0.943$\rightarrow$0.943$\rightarrow$0.943$\rightarrow$0.943$\rightarrow$0.950$\rightarrow$0.950$\rightarrow$0.950$\rightarrow$0.948  & \multirow{3}{*}{0.983} \\ %
    & SA & 0.963$\rightarrow$0.998$\rightarrow$1.000$\rightarrow$1.000$\rightarrow$1.000$\rightarrow$1.000$\rightarrow$1.000$\rightarrow$1.000$\rightarrow$0.998$\rightarrow$1.000 & \\ 
    & VA & 0.998$\rightarrow$0.998$\rightarrow$1.000$\rightarrow$1.000$\rightarrow$1.000$\rightarrow$1.000$\rightarrow$1.000$\rightarrow$1.000$\rightarrow$1.000$\rightarrow$1.000 & \\
    \midrule   
    
    \multirow{3}{*}{\textbf{Llama 3.3 70B Instruct}}
    & RA & 0.830$\rightarrow$0.945$\rightarrow$0.933$\rightarrow$0.938$\rightarrow$0.948$\rightarrow$0.958$\rightarrow$0.963$\rightarrow$0.953$\rightarrow$0.963$\rightarrow$0.963 & \multirow{3}{*}{0.988} \\ %
    & SA & 0.920$\rightarrow$0.978$\rightarrow$0.990$\rightarrow$0.980$\rightarrow$0.978$\rightarrow$0.988$\rightarrow$0.998$\rightarrow$1.000$\rightarrow$1.000$\rightarrow$1.000 & \\ 
    & VA & 0.993$\rightarrow$0.993$\rightarrow$1.000$\rightarrow$1.000$\rightarrow$1.000$\rightarrow$1.000$\rightarrow$1.000$\rightarrow$1.000$\rightarrow$1.000$\rightarrow$1.000 & \\
    \midrule
    
    \multirow{3}{*}{\textbf{DeepSeek-V3.2-Exp}}
    & RA & 0.895$\rightarrow$0.983$\rightarrow$0.995$\rightarrow$0.998$\rightarrow$0.998$\rightarrow$0.995$\rightarrow$1.000$\rightarrow$0.998$\rightarrow$1.000$\rightarrow$0.998  & \multirow{3}{*}{0.998} \\ %
    & SA & 0.948$\rightarrow$0.993$\rightarrow$0.993$\rightarrow$0.993$\rightarrow$0.995$\rightarrow$0.998$\rightarrow$0.998$\rightarrow$0.998$\rightarrow$0.998$\rightarrow$0.998 & \\ 
    & VA & 0.990$\rightarrow$0.995$\rightarrow$1.000$\rightarrow$0.998$\rightarrow$1.000$\rightarrow$0.998$\rightarrow$1.000$\rightarrow$1.000$\rightarrow$1.000$\rightarrow$0.998 & \\
    
    \bottomrule
    \end{tabular}
\label{round=10}
\end{table*}

\begin{figure}[h!]
    \centering
    \begin{minipage}[c]{0.7\textwidth}
        \centering
        \includegraphics[width=\textwidth]{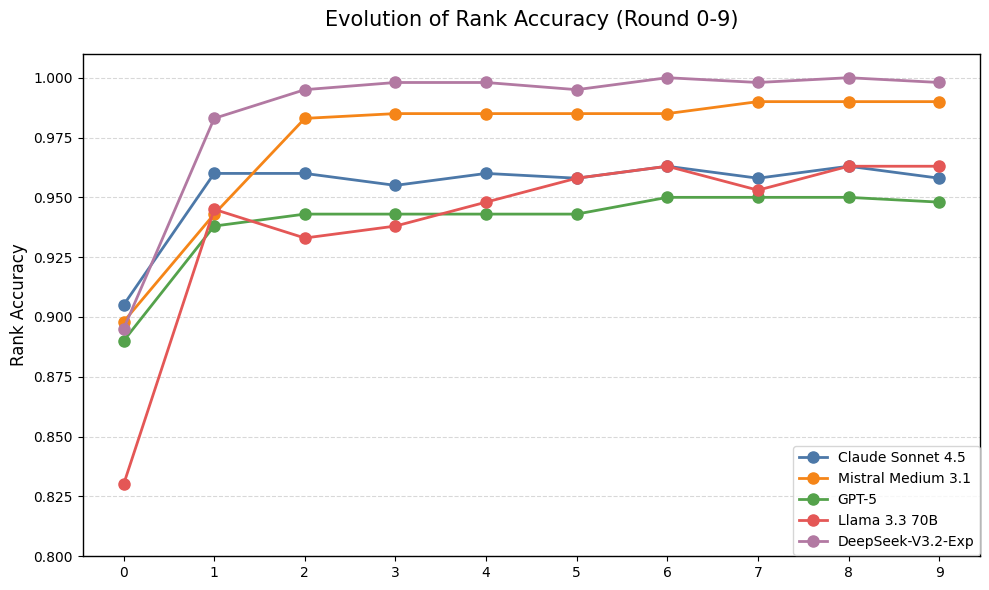}
    \end{minipage}
    \hfill
    \begin{minipage}[c]{0.7\textwidth}
        \centering
        \includegraphics[width=\textwidth]{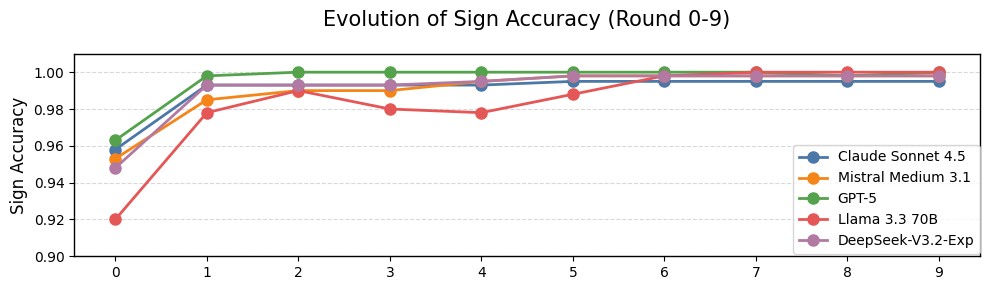}
        \vspace{0.5em} %
        \includegraphics[width=\textwidth]{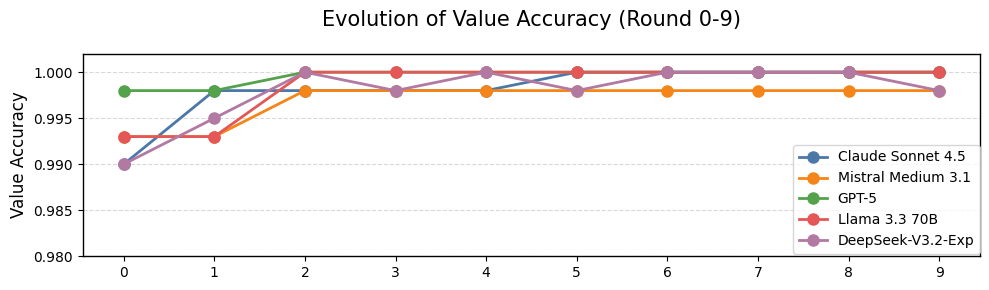}
    \end{minipage}
    
    \caption{The Evolution of \textit{Rank Accuracy (RA), Sign Accuracy (SA), and Value Accuracy (VA)} across 10 rounds for different LLMs.}
    \label{round=10plot}
\end{figure}

\textbf{Different Baseline Narratives} We wonder how much the various quality of baseline narratives influences the final results. Therefore, we use the best and the worst baseline narratives to add two extra experiments. According to the no-baseline setting results, the best baseline narratives are from DeepSeek-V3.2-Exp, and the worst baseline narratives are from Llama 3.3 70B Instruct. We derive three key insights from the results (Table~\ref{4additional}). First, across all LLMs, performance under the best baseline narratives consistently exceeds that under the worst baseline narratives. Second, DeepSeek-V3.2-Exp achieves the first and only reported 100\% faithfulness across all experiments in this paper. Third, different LLMs exhibit distinct patterns to baseline quality. Claude Sonnet 4.5 and Mistral Medium 3.1 show negligible variation across the three baseline settings, with round-2 accuracy differences within 0.003. GPT-5 demonstrates minimal difference between the original and best-baseline settings; however, its performance declines noticeably under the worst-baseline condition. In contrast, Llama 3.3 70B Instruct and DeepSeek-V3.2-Exp follow the anticipated trend more clearly: their round-2 accuracy improves significantly when provided with the best baseline narratives and decreases when initialized with the lowest-quality narratives.

In summary, we have three main conclusions on those additional experiments. First, we validate that the agentic system remains effective with the increase of the number of features included. However, the overall improvement effect weakens compared to the setting with fewer features, as reflected by the lower final-round accuracy. Second, under different baseline settings, the quality of the initial baseline narratives plays an important role. In general, better baseline narratives lead to stronger final performance. Although some LLMs demonstrate no obvious difference under different settings (Claude Sonnet 4.5), using the worst baseline narratives consistently decreases the final accuracy, and using the best narratives normally increases the effect or at least causes no harm. Third, increasing the maximum number of refinement rounds does not necessarily lead to higher final accuracy for most advanced LLMs, with the exception of Llama 3.3 70B Instruct, which continues to benefit from additional refinement rounds.

\begin{table*}[t]
\centering
\caption{The round=10 experiments' problem categorization (\textit{Critic-Rule Design})}
\label{round=10table}
\footnotesize %
\begin{tabular}{l c c c}
\toprule
\textbf{LLMs} & \textbf{No. Unf. Narr. (round 9)} & \textbf{C1: Confusing Faithfulness Instruction}  & \textbf{C2: Extraction Mistake} \\
\midrule

\textbf{Claude Sonnet 4.5} & 8 & 2& 6\\
\textbf{Mistral Medium 3.1} & 4 & $/$ & 4\\
\textbf{GPT-5} &9 & 2& 7\\
\textbf{Llama 3.3 70B Instruct} &5 & $/$& 5\\
\textbf{DeepSeek-V3.2-Exp} & 1 & $/$ & 1\\
\bottomrule
\end{tabular}

\end{table*}

\begin{table*}[t]
\centering
\caption{The additional experiments' results across different LLMs (\textit{Critic-Rule Design})}
\footnotesize %
\setlength{\tabcolsep}{5pt}
\begin{tabular}{l l ccc ccc c}
\toprule
\multirow{2}{*}{\textbf{LLMs}} & \multirow{2}{*}{\textbf{Experiments}} & \multicolumn{3}{c}{\textbf{Faithfulness}} & \multirow{2}{*}{\textbf{Overall (round 2)}} \\
\cmidrule(lr){3-5}
& & \textbf{RA} & \textbf{SA} & \textbf{VA} & \\
\midrule

\multirow{4}{*}{\textbf{Claude Sonnet 4.5}} 
& Original Exp.& 0.905$\rightarrow$0.960$\rightarrow$0.960 & 0.957$\rightarrow$0.993$\rightarrow$0.997 & 0.990$\rightarrow$0.995$\rightarrow$0.997 & 0.985 \\
& Feature=8    & 0.750$\rightarrow$0.924$\rightarrow$0.939 & 0.959$\rightarrow$0.991$\rightarrow$0.993 & 0.993$\rightarrow$0.994$\rightarrow$0.998 & 0.976 \\
& No Baseline  & 0.895$\rightarrow$0.970$\rightarrow$0.970 & 0.983$\rightarrow$0.990$\rightarrow$0.995 & 0.978$\rightarrow$0.990$\rightarrow$0.995 & 0.987 \\
& Best Base.   & 0.940$\rightarrow$0.965$\rightarrow$0.975 & 0.958$\rightarrow$0.983$\rightarrow$0.990 & 0.993$\rightarrow$0.998$\rightarrow$1.000 & 0.988 \\
& Worst Base.  & 0.775$\rightarrow$0.965$\rightarrow$0.970 & 0.968$\rightarrow$0.988$\rightarrow$0.993 & 0.985$\rightarrow$0.993$\rightarrow$0.998 & 0.987 \\
\midrule

\multirow{4}{*}{\textbf{Mistral Medium 3.1}} 
& Original Exp.& 0.907$\rightarrow$0.955$\rightarrow$0.980 & 0.952$\rightarrow$0.985$\rightarrow$0.990 & 0.990$\rightarrow$0.995$\rightarrow$0.997 & 0.989 \\
& Feature=8    & 0.729$\rightarrow$0.768$\rightarrow$0.808 & 0.940$\rightarrow$0.963$\rightarrow$0.969 & 0.995$\rightarrow$0.983$\rightarrow$0.996 & 0.924 \\
& No Baseline  & 0.770$\rightarrow$0.875$\rightarrow$0.925 & 0.928$\rightarrow$0.973$\rightarrow$0.978 & 0.983$\rightarrow$0.995$\rightarrow$0.995 & 0.966 \\
& Best Base.   & 0.945$\rightarrow$0.958$\rightarrow$0.973 & 0.968$\rightarrow$0.990$\rightarrow$0.998 & 0.990$\rightarrow$0.995$\rightarrow$0.998 & 0.989 \\
& Worst Base.  & 0.725$\rightarrow$0.868$\rightarrow$0.970 & 0.950$\rightarrow$0.975$\rightarrow$0.993 & 0.983$\rightarrow$0.995$\rightarrow$0.995 & 0.986 \\
\midrule

\multirow{4}{*}{\textbf{GPT-5}} 
& Original Exp.& 0.885$\rightarrow$0.957$\rightarrow$0.968 & 0.963$\rightarrow$0.997$\rightarrow$1.000 & 0.993$\rightarrow$0.997$\rightarrow$1.000 & 0.989 \\
& Feature=8    & 0.740$\rightarrow$0.816$\rightarrow$0.829 & 0.949$\rightarrow$0.978$\rightarrow$0.991 & 0.995$\rightarrow$0.995$\rightarrow$0.999 & 0.940 \\
& No Baseline  & 0.930$\rightarrow$0.953$\rightarrow$0.953 & 0.980$\rightarrow$1.000$\rightarrow$1.000 & 0.988$\rightarrow$1.000$\rightarrow$1.000 & 0.984 \\
& Best Base.   & 0.925$\rightarrow$0.955$\rightarrow$0.960 & 0.975$\rightarrow$1.000$\rightarrow$1.000 & 0.993$\rightarrow$1.000$\rightarrow$1.000 & 0.987 \\
& Worst Base.  & 0.743$\rightarrow$0.860$\rightarrow$0.865 & 0.978$\rightarrow$0.993$\rightarrow$1.000 & 0.990$\rightarrow$1.000$\rightarrow$1.000 & 0.955 \\
\midrule

\multirow{4}{*}{\textbf{Llama 3.3 70B Instruct}} 
& Original Exp.& 0.847$\rightarrow$0.912$\rightarrow$0.930 & 0.940$\rightarrow$0.970$\rightarrow$0.990 & 0.993$\rightarrow$0.993$\rightarrow$1.000 & 0.973 \\
& Feature=8    & 0.656$\rightarrow$0.764$\rightarrow$0.804 & 0.940$\rightarrow$0.950$\rightarrow$0.968 & 0.993$\rightarrow$0.998$\rightarrow$0.999 & 0.923 \\
& No Baseline  & 0.693$\rightarrow$0.823$\rightarrow$0.898 & 0.920$\rightarrow$0.935$\rightarrow$0.985 & 0.993$\rightarrow$1.000$\rightarrow$1.000 & 0.961 \\
& Best Base.   & 0.915$\rightarrow$0.935$\rightarrow$0.958 & 0.953$\rightarrow$0.950$\rightarrow$0.990 & 0.990$\rightarrow$1.000$\rightarrow$0.998 & 0.982 \\
& Worst Base.  & 0.725$\rightarrow$0.843$\rightarrow$0.888 & 0.943$\rightarrow$0.963$\rightarrow$0.968 & 0.995$\rightarrow$0.998$\rightarrow$1.000 & 0.952 \\
\midrule

\multirow{4}{*}{\textbf{DeepSeek-V3.2-Exp}} 
& Original Exp.& 0.912$\rightarrow$0.970$\rightarrow$0.978 & 0.960$\rightarrow$0.980$\rightarrow$0.980 & 0.992$\rightarrow$0.997$\rightarrow$1.000 & 0.986 \\
& Feature=8    & 0.783$\rightarrow$0.938$\rightarrow$0.968 & 0.946$\rightarrow$0.978$\rightarrow$0.989 & 0.990$\rightarrow$0.995$\rightarrow$0.999 & 0.985 \\
& No Baseline  & 0.980$\rightarrow$1.000$\rightarrow$1.000 & 0.968$\rightarrow$0.990$\rightarrow$0.998 & 0.990$\rightarrow$0.998$\rightarrow$1.000 & 0.999 \\
& Best Base.   & 0.978$\rightarrow$1.000$\rightarrow$1.000 & 0.968$\rightarrow$0.990$\rightarrow$1.000 & 0.988$\rightarrow$0.998$\rightarrow$1.000 & \colorbox{green!20}{\textbf{1.000}} \\
& Worst Base.  & 0.813$\rightarrow$0.973$\rightarrow$0.983 & 0.983$\rightarrow$0.995$\rightarrow$0.995 & 0.990$\rightarrow$0.998$\rightarrow$0.998 & 0.992 \\

\bottomrule
\end{tabular}
\label{4additional}

\end{table*}

\subsection{Coherence Analysis}\label{coanalysis}

To understand the suggestion provided by the \textit{Coherence Agent} and their influence on the narrative, we, again, use the outputs from Claude Sonnet 4.5-based \textit{Coherent-Rule Design} (original experiment) to conduct a sampled, qualitative analysis. In general, the \textit{Coherence Agent} either reports no coherence issues or proposes revisions suggestions using the four templates we allow: ``Change \_\_\_ to \_\_\_”, ``Insert \_\_\_ before \_\_\_”, ``Delete \_\_\_”, ``Reorder \_\_\_ after \_\_\_”. After a close examination, we observe these suggestions mostly cover the following aspects: transition re-wordings, sentence restructuring or rephrasing, adjustments to feature ordering, influence direction, or value modifications, etc. We show each a real example as follows:

\textbf{a) No coherence issue:} The \textit{Coherence Agent} gives a final judgment: ``no coherence issue" with an explanation. Accordingly, the \textit{Narrator} will not change. An example:

\begin{tcolorbox}[colback=white!5, colframe=darkbluegrey, title=\textbf{\{No Coherence Issue\} Student: index 9, round 2 feedback}]

After examining the narrative for coherence issues, I find:

\colorbox{green!20}{no coherence issue}

The narrative demonstrates strong logical coherence throughout:

1.\textbf{Clear structure}: Opens with the prediction outcome, then systematically ...

2.\textbf{Smooth transitions}: Uses appropriate connectives to guide ...

3.\textbf{Logical flow}: Progresses from most influential factor to less ...

4.\textbf{Unified explanation}: Each paragraph builds on the previous one ...

5.\textbf{Effective conclusion}: The final sentence synthesizes the analysis by ...

6.\textbf{Consistent reasoning}: Counterintuitive findings (family support) are acknowledged ...

The text maintains topical unity, avoids abrupt transitions...

\end{tcolorbox}

\textbf{b) Transitional wordings}: The \textit{Coherence Agent} gives suggestions to add or change the transitional wordings or phrases. An example is shown below. In the next round, the \textit{Narrator} changes the narrative accordingly. Although this kind of suggestion is nuanced and minor, it indeed improves the flow of the whole narrative. 

\begin{tcolorbox}[colback=white!5, colframe=darkbluegrey, title=\textbf{\{Transitional Wordings\} Diabetes: index 9, round 0 feedback}]

...

2) Insert: \colorbox{green!20}{``In contrast," } before ``The Diabetes Pedigree Function score of 0.263"  

\textbf{Justification}: This transition phrase explicitly signals the shift from a risk-increasing factor to a risk-decreasing factor, improving the logical flow...

3) Change: ``While the patient's age of 29 was younger than average, it still contributed a small increase in risk"  To: ``Although the patient's age of 29 was younger than average, it \colorbox{green!20}{paradoxically} contributed a small increase in risk" 

\textbf{Justification}: The word ``paradoxically" makes the counterintuitive relationship more explicit, helping readers ...

...

\end{tcolorbox}

\textbf{c) Sentence restructure}: The \textit{Coherence Agent} suggests restructuring or rephrasing sentences to make the narrative more understandable, clear, and accessible. For example, it recommends splitting a long sentence containing multiple clauses into two shorter sentences:

\begin{tcolorbox}[colback=white!5, colframe=darkbluegrey, title=\textbf{\{Sentence Restructure\} Student: index 12, round 1 feedback}]

...

3. Change ``which actually worked against their prediction slightly, as the model expected some absences based on typical patterns and their perfect attendance may signal other unmeasured factors" to ``which surprisingly worked against their prediction. The model associates zero absences with other unmeasured risk factors, as typical passing students show some absences"

\textbf{Justification}: The original sentence is convoluted with multiple clauses.
\colorbox{green!20}{Breaking it} into clearer statements and reversing the logic (explaining what the model expects for passing students) makes the counterintuitive finding more comprehensible.

...

\end{tcolorbox}

\textbf{d) Suggestions on changing the rank/sign/value
}: The\textit{ Coherence Agent} suggests changing the feature rank, influence direction, or feature value to make the whole narrative more plausible, logical, and sensible. For example, it suggests grouping those features with positive influences together. However, these suggestions may violate the information from the SHAP table. This is the primary source of degraded faithfulness in the revised narratives.

\begin{tcolorbox}[colback=white!5, colframe=darkbluegrey, title=\textbf{\{Suggestions on changing the rank/sign/value\} Credit: index 6, round 2 feedback}]

...

1. Reorder the sentence ``However, the presence of other installment plans..." to \colorbox{green!20}{appear after}  all positive factors are discussed.

\textbf{Justification}: The current placement interrupts the flow of positive factors.
The narrative jumps from checking account status (positive) to installment plans (negative) back to employment duration (positive), creating a disjointed reading experience. Grouping all positive factors together...

...

\end{tcolorbox}

To summarize, after adding the \textit{Coherence Agent}, the system includes a dedicated module for assessing the overall linguistic quality of the narrative. It identifies coherence issues that are particularly effective at refining subtle aspects of sentence wording and structure that the \textit{Narrator} may struggle to detect and improve. However, as its sole focus is on coherence, it may generate suggestions that conflict with the SHAP table, thereby guiding the \textit{Narrator} toward revisions that compromise faithfulness. Overall, while the inclusion of the \textit{Coherence Agent} improves linguistic quality, this often comes at the expense of faithfulness. The interplay between these two evaluation dimensions, and how the agents coordinate within the system, is considered to be a direction for future research.

\section{Conclusion and Outlook}
In this work, we explore an agentic approach to generating and improving SHAP-based narratives. We propose a multi-agent framework in which the \textit{Narrator} generates and revises narratives in response to feedback from other agents. The \textit{Faithful Evaluator} and \textit{Faithful Critic} provide feedback in a quantitative approach, based on three faithfulness metrics: \textit{Rank Accuracy}, \textit{Sign Accuracy}, and \textit{Value Accuracy}, whereas the \textit{Coherence Agent} gives feedback on the linguistic coherence quality in a qualitative way. We design five different system variants and test their effectiveness using five LLMs. The \textit{Basic Design}, the \textit{Critic Design}, and the \textit{Critic-Rule Design} are validated to be effective in improving the faithfulness level across all five LLMs; however, no single design consistently outperforms others across all LLMs. The comparable performance of the \textit{Critic Design} and \textit{Critic-Rule Design} suggests that the rule-based \textit{Faithful Critic} is a better and more resource-efficient solution among the two. The Claude Sonnet 4.5-based \textit{Basic Design} ranks first among the 25 experimental combinations, yet does not achieve the reported 100\% faithfulness. A closer examination reveals that the extraction mistakes by a single LLM are the main recurring issue for all experiments. We hence propose an ensemble method that introduces all five LLMs as independent \textit{Faithful Evaluators}, using their consensus as the final judgment. This ensemble method mitigates the extraction mistakes but does not eliminate them entirely, as such errors are largely unavoidable due to inherent linguistic ambiguity and semantic nuance. Beyond this, we conduct additional experiments on several key concepts to further investigate the conditions required for an effective agentic system. We validate that the agentic system remains effective with the increase of the number of features included in one narrative. Increasing the round of iteration not necessarily improve the final faithfulness quality; in contrast, the quality of the baseline narratives does indeed. By replacing the baseline narratives with higher-quality ones, the DeepSeek-V3.2-Exp-based \textit{Critic-Rule Design} achieves, for the first time, a reported 100\% faithfulness. On the other hand, the introduction of the \textit{Coherence Agent} generally improves the narrative's linguistic quality after a qualitative, manual sample check; however, this inevitably harms the faithfulness of SHAP narratives.     

This work identifies some drawbacks and outlines future work. First, we haven't found a solution for the extraction mistakes. Semantic nuances pose inherent difficulties for a single LLM to consistently extract the correct information, even for DeepSeek-V3.2-Exp, the most performant LLM on this task. The ensemble method we propose does not fully resolve the issue, in part because some LLMs, such as Llama 3.3 70B Instruct, perform poorly on this task, making them poor candidates and leading to incorrect consensus. Moreover, an LLM may yield different extractions for the same narrative across attempts. Future work may therefore focus on either filtering out unreliable candidates from the majority voting or using DeepSeek-V3.2-Exp multiple times to encourage more robust extraction. This also implicates the possibility of integrating the different LLMs into one agent system. Given the round-0 narrative's significance, it is important to deploy an advanced LLM as the \textit{Narrator}, such as DeepSeek-V3.2-Exp or GPT-5. Another drawback is the qualitative analysis of the coherence feedback. So far, there is no effective quantitative metric of coherence that can be integrated into the iteration cycle. Due to the workload, the manual analysis of coherence feedback is also time-consuming. Future work can incorporate LLM-as-a-judge as one downstream agent to mark and filter the coherence problem, thereby simplifying qualitative analysis. Moreover, given the conflicting effects of the \textit{Coherence Agent} and the faithfulness-related agents, exploring their interplay and identifying a means to facilitate cooperation should also be an important topic. Last, since comprehensibility and controllability are important factors in explanations, future work can include such modules so that people can interact with the system to generate user-personalized narratives based on their own preferences and cognitive level.

\appendix
\section{Appendices}

\subsection{Prompts for each agent}\label{prompts}

\begin{tcolorbox}[colback=blue!5, colframe=darkbluegrey, title=\textbf{Base Prompt (Structure)}]

\textbf{Explanation goal:}
Description of goal: generate narrative.

\textbf{Summary SHAP methodology:}
Description of SHAP.

\textbf{Dataset context:} 
Descriptions of dataset/target/task.

\textbf{SHAP table:}

\begin{tabularx}{1.0\textwidth} { 
  | >{\raggedright\arraybackslash}X 
  | >{\centering\arraybackslash}X 
  | >{\centering\arraybackslash}X 
  | >{\centering\arraybackslash}X 
  | >{\centering\arraybackslash}X | }
 \hline
 Feature & SHAP & Feat.val. & Feat.avg. & Feat.desc.\\
 \hline
 Goals  & 0.135  & 2  & 1.5 & ... \\
\hline
 Attempts  & -0.120  & 12  & 12.5 & ...\\
\hline
 ...  & ...  & ...  & ... & ...\\
\hline
\end{tabularx}
\bigskip

\textbf{Result string: }Short string stating the probability predicted for class 1.

\textbf{Format related rules:} Several rules about the format of the narrative.

\textbf{Content related rules:} Several rules about what to emphasize in the narrative.

\label{baseprompt}
\end{tcolorbox}

\begin{tcolorbox}[colback=blue!5, colframe=darkbluegrey, title=\textbf{Narrator prompt}]
\textbf{Context:}   

You are a helpful agent who writes model explanations (narratives) based on SHAP values. 

Revise your previous narrative strictly according to the initial task and all given feedback. 
            
This is your initial task: \{initial\_prompt\}.
\bigskip 

\textbf{Input text:}

The following is the feedback. 

====================

This is your previous answer:\{last\_narrative\}.

This is the faithfulness-issue feedback:\{faithful\_feedback\}.

This is the coherence-issue feedback:\{coherence\_feedback\}.

====================
\bigskip 

\textbf{Output Structure:}

The narrative MUST comply with all format related rules and content related rules from the initial task.
\bigskip 

\textbf{Guidelines:}

1) Do not modify the part of the narrative that isn't mentioned in the feedback. 

2) You MUST return the narrative only. DO NOT chitchat.

\label{Narrator}
\end{tcolorbox}

\begin{tcolorbox}[colback=blue!5, colframe=darkbluegrey, title=\textbf{Faithful Evaluator prompt}]

\textbf{Context:}

        An LLM was used to create a narrative to explain and interpret a prediction made by another smaller classifier model. 
        The LLM was given an explanation of the classifier task, the training data, and provided with the exact names of all the features and their meaning. 
        Most importantly, the LLM was provided with a table that contains the feature values of that particular instance, the average feature values and their SHAP values which are a numeric measure of their importance. 

        \bigskip
        
        You are an helpful agent tasked with improving the narrative. 
        To do so, you should extract some information about all the features that were mentioned in the narrative that will be given below.
        \bigskip
        
        Here is some general info for you:
        
        Dataset description: \{self.ds\_info["dataset\_description"]\}.
        
        Target description: \{self.ds\_info["target\_description"]\}.
        
        Task description: \{self.ds\_info["task\_description"]\}.
        
        Feature descriptions: \{self.feature\_desc[["feature\_name","feature\_desc"]].to\_string(index = False)\}.
                \bigskip
        
        \textbf{Input text: }
        
        The following is the given narrative.
        
        ====================
        
        \{narrative\}
        
        ====================
 \bigskip
 
        \textbf{Output Structure:}
        
        Provide your answer as a python dictionary with the keys as the feature names. 
        The values corresponding to the feature name keys are dictionaries themselves that contain the following inner keys: 
         \bigskip
         
        1) "rank:" indicating the order of absolute importance of the feature starting from 0.
        
        2) "sign": the sign of whether the feature contributed towards target value 1 or against it (either +1 or -1 for sign value).
        
        3) "value": if the value of the feature is mentioned in a way that you can put an exact number on, add it. Only return numeric values here.
        If the description of the value is qualitative such as "many" or "often" and not mentioning an exact value, return "None" for its value.
        
        4) "assumption": give a short but complete 1 sentence summary of what the assumption is in the story for this feature.
        Provide this assumption as a general statement that could be fact checked later and that does not require this narrative as context.
        If no reason or suggestion is made in the story do not make something up and just return string 'None'.
                \bigskip
        
        \textbf{Guidelines:}
        
        1) Make sure that both the "rank", "sign", "value" and "assumption" keys and their values are always present in the inner dictionaries.
        
        2) Make sure that the "rank" key is sorted from 0 to an increasing value in the dictionary. The first element cannot have any other rank than 0.
        
        3) Make sure to use the exact names of the features as provided in the Feature descriptions, including capitalization. 
        
        4) Just provide the python dictionary as a string and add nothing else to the answer.

\label{Evaluator}
\end{tcolorbox}

\begin{tcolorbox}[colback=blue!5, colframe=darkbluegrey, title=\textbf{Faithful Critic prompt}]
\textbf{Context: }  

You are a critic tasked with providing instructions on how to improve a narrative. 
            To do so, you are given feedback on the narrative, and you should summarize it clearly and concisely.

\bigskip 

\textbf{Input text:}

The following is the feedback. 

====================

\{combined\_feedback\}

====================

\bigskip 

\textbf{Output Structure:}

Free format (no strict structure required).

\bigskip 

\textbf{Guidelines:}

When you summarize, make sure to include all feedback; do not lose any information from the feedback provided.

\label{Critic}
\end{tcolorbox}

\begin{tcolorbox}[colback=blue!5, colframe=darkbluegrey, title=\textbf{Coherence Agent prompt}]

\textbf{Context:}

                You are a critic tasked with providing revision instructions to improve the coherence quality of the given narrative.
        You should first examine if coherence-related issues exist in the given narrative and then output revision instructions.
        
\bigskip

        \textbf{Definition of coherence:}
        
        Coherence refers to the overall quality of how sentences work together in a text. A coherent text is well-structured, logically organized, and builds a unified body of information on its topic. 
        It should present information that flows smoothly, avoiding abrupt transitions or disjoint statements.

\bigskip

\textbf{Input text:}

        The following is the given narrative. 
        
        ====================
        
        \{narrative\}
        
        ====================
        
\bigskip

    \textbf{Output Structure:}
    
        Your output MUST include:
        
        1) 1) Explicit revision commands written in a clear, standardized format, such as: “Change \_  to \_, Insert \_ before \_, Delete \_, Reorder \_ after \_”, etc.
        
        2) A concise explanation following each command that briefly justifies the change.

\bigskip

        \textbf{Guidelines:}
        
        1) If there are no coherence issues, reply only: no coherence issue.
        
        2) Focus on meaningful coherence improvements. Avoid nitpicking or unnecessary edits. 

\label{Coherence}
\end{tcolorbox}

\subsection{LLM Extraction Instability}\label{llmexapp}
It is also worth noting that extraction instability across different LLMs varies (Table~\ref{llmex}). As we already know from Table~\ref{claudetable}, Claude Sonnet 4.5 consistently extracts information, differing only once from the others in round-0 sign accuracy across five designs. It is hence the most robust and consistent LLM in this extraction task. GPT-5 and DeepSeek-V3.2-Exp, although possessing as good improvement ability as for Claude Sonnet 4.5, show big variances on the round-0 rank accuracy, whereas Llama 3.3 70B Instruct is the most unstable LLM in this task. Therefore, when comparing the effects of improvement, some variance tolerance should be considered.

\begin{table}[h!]
\centering
\footnotesize
\caption{LLM extraction instability on round-0 rank, sign and value accuracy. The mean is computed on the corresponding accuracy across five agentic system designs.}
\begin{tabular}{lccccc}
\toprule
\textbf{LLMs} & \textbf{Faithfulness} & \textbf{Mean} & \textbf{Min} & \textbf{Max} & \textbf{Std Dev.} \\ 
\midrule
\multirow{3}{*}{Claude Sonnet 4.5} & RA & 0.905 & 0.905 & 0.905 & 0.000 \\
& SA & 0.958 & 0.957 & 0.960 & 0.001 \\
& VA & 0.990 & 0.990 & 0.990 & 0.000 \\
\midrule
\multirow{3}{*}{Mistral Medium 3.1} & RA & 0.900 & 0.897 & 0.907 & 0.004 \\
& SA & 0.952 & 0.950 & 0.952 & 0.001 \\
& VA & 0.990 & 0.988 & 0.993 & 0.002 \\
\midrule
\multirow{3}{*}{GPT-5} & RA & 0.893 & 0.885 & 0.902 & 0.007 \\
& SA & 0.964 & 0.963 & 0.965 & 0.001 \\
& VA & 0.993 & 0.993 & 0.993 & 0.000 \\
\midrule
\multirow{3}{*}{Llama 3.3 70B Instruct} & RA & 0.863 & 0.847 & 0.875 & 0.010 \\
& SA & 0.947 & 0.940 & 0.953 & 0.006 \\
& VA & 0.993 & 0.993 & 0.993 & 0.000 \\
\midrule
\multirow{3}{*}{DeepSeek-V3.2-Exp} & RA & 0.911 & 0.905 & 0.922 & 0.006 \\
& SA & 0.954 & 0.950 & 0.960 & 0.005 \\
& VA & 0.993 & 0.992 & 0.995 & 0.001 \\
\bottomrule
\end{tabular}
\label{llmex}
\end{table}

\subsection{Best Agentic System Design Selection} \label{benchmarkselection}

\begin{table*}[h!]
\centering
\caption{The number of occurrences of the best design for each LLM}
\footnotesize
    \setlength{\tabcolsep}{4pt}
    \begin{tabular}{l l c c}
    \toprule
    \multirow{2}{*}{\textbf{LLMs}} & \multirow{2}{*}{\textbf{Agentic System Designs}} & \textbf{Average (round 2)} & \multirow{2}{*}{\textbf{Count}}\\
    & & \textbf{(o)$\vert$(e)} & \\
    \midrule
    
    \multirow{5}{*}{\textbf{\makecell[l]{Claude Sonnet 4.5}}}
    & Basic Design
    & \textcolor{teal}{0.996}$\vert$\textcolor{teal}{0.999}
    & 2\\
    & Critic Design
    & 0.983$\vert$0.988 
    & 0\\
    & Critic-Rule Design
    & 0.985$\vert$0.988 
    & 0\\
    & Coherent Design
    & 0.946$\vert$0.954 
    & 0\\
    & Coherent-Rule Design
    & 0.953$\vert$0.962 
    & 0\\
    \midrule
    
    \multirow{5}{*}{\textbf{\makecell[l]{Mistral Medium 3.1}}}
    & Basic Design
    & 0.951$\vert$0.985 
    & 0\\
    & Critic Design
    & \textcolor{teal}{0.988}$\vert$\textcolor{teal}{0.996} 
    & 2\\
    & Critic-Rule Design
    & \textcolor{teal}{0.989}$\vert$0.984
    & 1\\
    & Coherent Design
    & 0.907$\vert$0.918 
    & 0\\
    & Coherent-Rule Design
    & 0.916$\vert$0.925 
    & 0\\
    \midrule
    
    \multirow{5}{*}{\textbf{\makecell[l]{GPT-5}}}
    & Basic Design
    & \textcolor{teal}{0.990}$\vert$\textcolor{teal}{0.993} 
    & 2\\
    & Critic Design
    & 0.982$\vert$0.988 
    & 0\\
    & Critic-Rule Design
    & \textcolor{teal}{0.989}$\vert$0.988
    & 1\\
    & Coherent Design
    & 0.954$\vert$0.968 
    & 0\\
    & Coherent-Rule Design
    & 0.953$\vert$0.978 
    & 0\\
    \midrule
    
    \multirow{5}{*}{\textbf{\makecell[l]{Llama 3.3 70B Instruct}}}
    & Basic Design
    & 0.946$\vert$0.970 
    & 0\\
    & Critic Design
    & 0.968$\vert$\textcolor{teal}{0.986} 
    & 1\\
    & Critic-Rule Design
    & \textcolor{teal}{0.973}$\vert$\textcolor{teal}{0.987} 
    & 2\\
    & Coherent Design
    & 0.927$\vert$0.953 
    & 0\\
    & Coherent-Rule Design
    & 0.890$\vert$0.946 
    & 0\\
    \midrule
    
    \multirow{5}{*}{\textbf{\makecell[l]{DeepSeek-V3.2-Exp}}}
    & Basic Design
    & 0.971$\vert$0.974 
    & 0\\
    & Critic Design
    & \textcolor{teal}{0.990}$\vert$\textcolor{teal}{0.985}
    & 2\\
    & Critic-Rule Design
    & 0.986$\vert$\textcolor{teal}{0.984} 
    & 1\\
    & Coherent Design
    & 0.967$\vert$0.956
    & 0\\
    & Coherent-Rule Design
    & 0.942$\vert$0.942 
    & 0\\
    
    \bottomrule
    \end{tabular}
    
    \footnotesize{To account for negligible variance, differences within 0.001 are treated as a tie, with both credited as joint winners.}
\label{benchmarkchoice}
\end{table*}

\subsection{LLM Selection and Costs} \label{LLMmodelselection}

\begin{table}[h!]
\centering
\footnotesize
\caption{The comparison of the five LLMs. The cost reported in each cell represents the total expense incurred by running all five agentic designs on 100 instances, with up to three refinement iterations.}
\begin{tabular}{llcccc}
\hline
\textbf{Model} & \textbf{Model (check point)} & \textbf{Geograph} & \textbf{Open source} & \textbf{Cost in total (Euro) (o) } & \textbf{Cost in total (Euro) (e)} \\
\hline
Claude Sonnet 4.5& claude-sonnet-4-5-20250929 & US & No & 10.7 & 31.8 \\
Mistral Medium 3.1 & mistral-medium-2508  &  Europe & No& 1.7  & 4.5  \\
GPT-5& gpt-5-chat-latest   & US & No & 4.7  & 10.6 \\
Llama-3.3-70B & llama-3.3-70b-instruct & US & open-weight & 0.04  & 1.2  \\
Deepseek-V3.2-Exp & deepseek-v3.2-exp   &  China &  open-weight & 0.6  & 2.0  \\
\hline
\end{tabular}
\end{table}

\bigskip

\printbibliography

\end{document}